\documentclass[11pt]{article}
\usepackage[letterpaper,margin=2.5cm]{geometry}
\usepackage{times}
\usepackage{amsmath}
\usepackage{amssymb}
\usepackage{amsthm}
\usepackage{graphicx}
\usepackage{xcolor}                  %
\usepackage{booktabs}
\usepackage{array}
\usepackage{rotating}                %
\usepackage[numbers,super,sort&compress]{natbib}
\usepackage[labelfont=bf,font=small]{caption}
\usepackage[hidelinks]{hyperref}
\newcommand{\val}[2]{\ensuremath{{\mathcode`\,="002C #1}}}

\newcolumntype{Q}{>{\raggedright\arraybackslash}p{0.19\textwidth}}
\newcolumntype{R}{>{\raggedright\arraybackslash}p{0.25\textwidth}}
\newcolumntype{S}{>{\raggedright\arraybackslash}p{0.3\textwidth}}
\newcolumntype{T}{>{\raggedright\arraybackslash}p{0.18\textwidth}}

\theoremstyle{plain}
\newtheorem{proposition}{Proposition}
\newtheorem{corollary}{Corollary}
\theoremstyle{remark}
\newtheorem{remark}{Remark}

\title{Which population does a benchmark ranking describe?}
\author{Justin Li\thanks{Correspondence: \href{mailto:jli89@tulane.edu}{jli89@tulane.edu}}\\[4pt]
  \small Center for Biomedical Informatics \& Genomics\\
  \small Tulane University School of Medicine, Tulane University\\
  \small New Orleans, LA, United States}
\date{}
\hypersetup{pdftitle={Which population does a benchmark ranking describe?},pdfauthor={Justin Li}}

\begin{document}
\maketitle

\begin{abstract}
Benchmark rankings describe the populations their construction retains. A label-free audit of ten benchmarks found
that seven excluded
\val{20}{card:ogbl-citation2@exclusion.W|pct}\textendash{}\val{99.97}{card:amazon-book@exclusion.W_pairs|pct}\% of
their evaluated units and five had restrictions their documentation did not fully describe. On WN18RR, expanding the
candidate set reversed the AUC ordering of ConvE and DistMult trained on the WordNet target graph. This was the only
reversal among \val{15}{count(excl0):E2/primary/*-*/mean/delta_O} comparisons under unadjusted per-pair intervals.
DistMult retained its lead in MRR, and the AUC reversal did not persist under degree reweighting.

When a benchmark keeps questions that few panel members solve, a member scores no higher on the questions kept with
its vote than on those kept with an outside model in its seat, provided the answers that voted are the answers
scored. On BigCodeBench, the penalty extended to another prompt format. Concentrating half the panel in one
model line also modestly disadvantaged that line's non-voting models relative to a more dispersed panel. On MMLU
questions retained by MMLU-Pro, sign changes beyond matched references clustered near chance performance and among
Llama-2 descendants. ScopeCard implements the fixed-rule audit, and the panel analyses come with reproducible
scripts.
\end{abstract}

Benchmark rankings decide which models are adopted, and they are read as statements about a task. A benchmark,
however, is the part of a population that its construction keeps. Knowledge-graph link-prediction benchmarks were
built from frequently mentioned entities\citep{pujara2017sparsity,mohamed2020popularity}; recommendation benchmarks
keep only users and items with enough interactions\citep{beel2019data,winter2025dataset}; retrieval collections keep
the topics with enough relevant judgments\citep{voorhees2022too}. Newer benchmarks are also built with models: they
drop the candidate questions that at least a threshold number of models in a panel answer
correctly\citep{wang2024mmlupro,zhuo2025bigcodebench,phan2025humanity}.
Construction rules thus restrict the population to which a model comparison applies, and when the filtering depends
on model outputs, its effect can also reach outputs and models that took no part in it. Benchmarks rarely report
either.

Two choices in construction decide which population a comparison describes. One is what a step removes: candidates
inside each query, as in link prediction and recommendation, or whole items and queries, as in question answering and
retrieval. The other is what decides the removal: a fixed eligibility rule, whose outcome depends on no model, or the
outputs of models. Under a fixed rule, a comparison on the benchmark describes the kept population under one metric,
and whether it carries over to the declared target depends on what the rule left out and on how the metric aggregates
over it. Under a filter that uses model outputs, the kept set also depends on which models voted, so a comparison can
penalise a model for having taken part, or for being related to a model that did. For either kind of rule, reading a
ranking beyond its benchmark raises two questions: what the release identifies about the population it kept, and what
evidence would show that the selection changed a comparison.

We give a diagnostic framework for both, and measure what each kind of rule does. First, for fixed rules, a per-query
identity and break-even threshold state when an unmeasured change on the excluded part could reverse a comparison;
when candidates are removed within queries, reciprocal rank and Hits@$k$ admit no such threshold, because regional
differences do not in general determine their ordering; and four regimes, decidable without labels, state what a
release identifies. An audit of ten widely reused
benchmarks measures the excluded share in the seven whose universe is available, states what the other three lack,
and tests whether the stated rules reproduce the releases. On WN18RR, the excluded synsets can
be recovered from WordNet, allowing us to test directly whether expanding the candidate set changes the model
ordering. ScopeCard turns this part
into a check that a benchmark's user can run, and a reproduction package recomputes the paper's numbers from public
inputs and stored model outputs, without retraining.

Second, for filters that use model outputs, an identity fixes the sign of a panel member's penalty for its own vote; archived
item-level runs of two benchmarks show how that penalty changed with the size of the panel, and on BigCodeBench that
it reached the members' outputs in another prompt format and that a panel concentrated in one model line put that
line's non-voting models at a small disadvantage. Third, on the original MMLU questions that MMLU-Pro's release
retained after its model filter and its later review, the order of models changes more than random and
difficulty-matched references give, mainly near the guessing floor and in the Llama-2 line that held half of the
panel's votes; no reversal is established among the filter models and the systems not derived from them. The
conclusion of a comparison therefore depends on the declared target and the
metric and, when model outputs built the benchmark, on which models built it.

\section*{Related work}

Evaluation research has asked whether benchmark conclusions carry over to the tasks that benchmarks stand
for\citep{liao2021are,raji2021ai}, how much test sets rebuilt by the original procedure change
results\citep{recht2019imagenet}, and how sampled candidates bias the metrics of
recommenders\citep{krichene2020sampled}, while use concentrates on ever fewer datasets\citep{koch2021reduced}.
Inside benchmarks, effects of popularity and degree have been
studied\citep{mohamed2020popularity,shomer2023toward,li2023evaluating}, and in recommendation the users that pruning
removed have been evaluated directly\citep{beel2019data,winter2025dataset}. For a drug--target interaction benchmark,
the negatives that its eligibility rule excludes have been measured: AUROC is lower there for every model, while the
order of the two trained models is unchanged\citep{li2026lowbias}. \citet{li2026constructed} gave a linear identity
and a break-even threshold for classification benchmarks whose negatives are constructed, with one fixed set of
positives; ranking evaluation differs, because every query has its own candidates, exclusion can remove candidates or
whole queries, and the metrics by which models are ranked do not average over candidates. Work on data leakage asks
whether a test set represents the data a model will meet in use\citep{bernett2024guiding} and provides splits that
reduce similarity between folds\citep{joeres2025datasail}. Such splits divide a dataset taken as given. For fixed rules
the question here comes earlier: what the dataset left out of the declared target, and whether that changes a
comparison. Generalizing a trial's effect to a target population raises the same question with covariates: weighting
the trial is unbiased when the moderators of the effect are measured and every unit could have been sampled, and
sensitivity analyses bound the bias that an omitted moderator leaves\citep{huang2022sensitivity}. A fixed eligibility
rule removes units with certainty, so no reweighting of the kept units reaches the excluded ones; the break-even
threshold here is stated on the unmeasured change between the regions, with the excluded share computed from the
declared universe.

Models are also used to build evaluation sets. Adversarial filtering removes examples that a model's representations
make easy\citep{lebras2020adversarial} or replaces wrong answers that a model can
detect\citep{zellers2018swag,zellers2019hellaswag}; adversarial collection keeps human-written examples that fool a
model\citep{nie2020adversarial}; and recent benchmarks drop the candidate questions that at least a threshold number
of models in a panel answer correctly\citep{wang2024mmlupro,zhuo2025bigcodebench,phan2025humanity}.
\citet{phang2021adversarially}
filtered natural-language-inference and question-answering sets against single adversary models. The filtered sets
were harder for every model, more so for stronger adversaries; the order of models differed between adversaries;
scores moved towards chance, which the authors offer as one reading of the reordering; and a model filtered against its
own pretrained base lost places. Items also function differently between model families: matched on total score,
descendants of Llama and Qwen on the Open LLM Leaderboard showed large differential item functioning on more than a
third of the items in five of six benchmarks, and held-out questions sorted by two features that predicted the
favoured family moved a near-tie between the families to gaps of six to eleven points in either direction on BBH and
MuSR\citep{lee2026dif}. \citet{zheng2026neartied} recomposed benchmarks from the items whose functioning
differs least between model families, chose the items with the responses of one half of the model owners and scored the
other half, and found that near-tied pairs of models from different families changed order more often than under random
subtests matched on item group and easiness, while rank correlation stayed high. On SWE-bench Verified, exact paired
tests separated none of the adjacent pairs among the top thirty submissions; rescoring them on the instances that some
but not all of the top twenty resolved widened the spread of scores and swapped three adjacent pairs, while still
separating none\citep{liu2026converged}. Scoring models on items chosen with
the same models' outputs is a form of circular analysis\citep{kriegeskorte2009circular}.
For filters that use model outputs, this paper adds an identity that fixes the sign of a member's penalty for its own
vote; the size of that penalty on random panels under the rules that released benchmarks use, and how it changed with
panel size; at a fixed panel size and approximately matched pass rates, its reach to outputs that never voted, in
another prompt format, and to the non-voting models of a line concentrated on the panel; and the order of models on
the questions that a released benchmark retained, read against references matched on subject, estimated difficulty
and answer key.

\section*{Results}

\subsection*{The population a comparison describes, and what a release identifies}

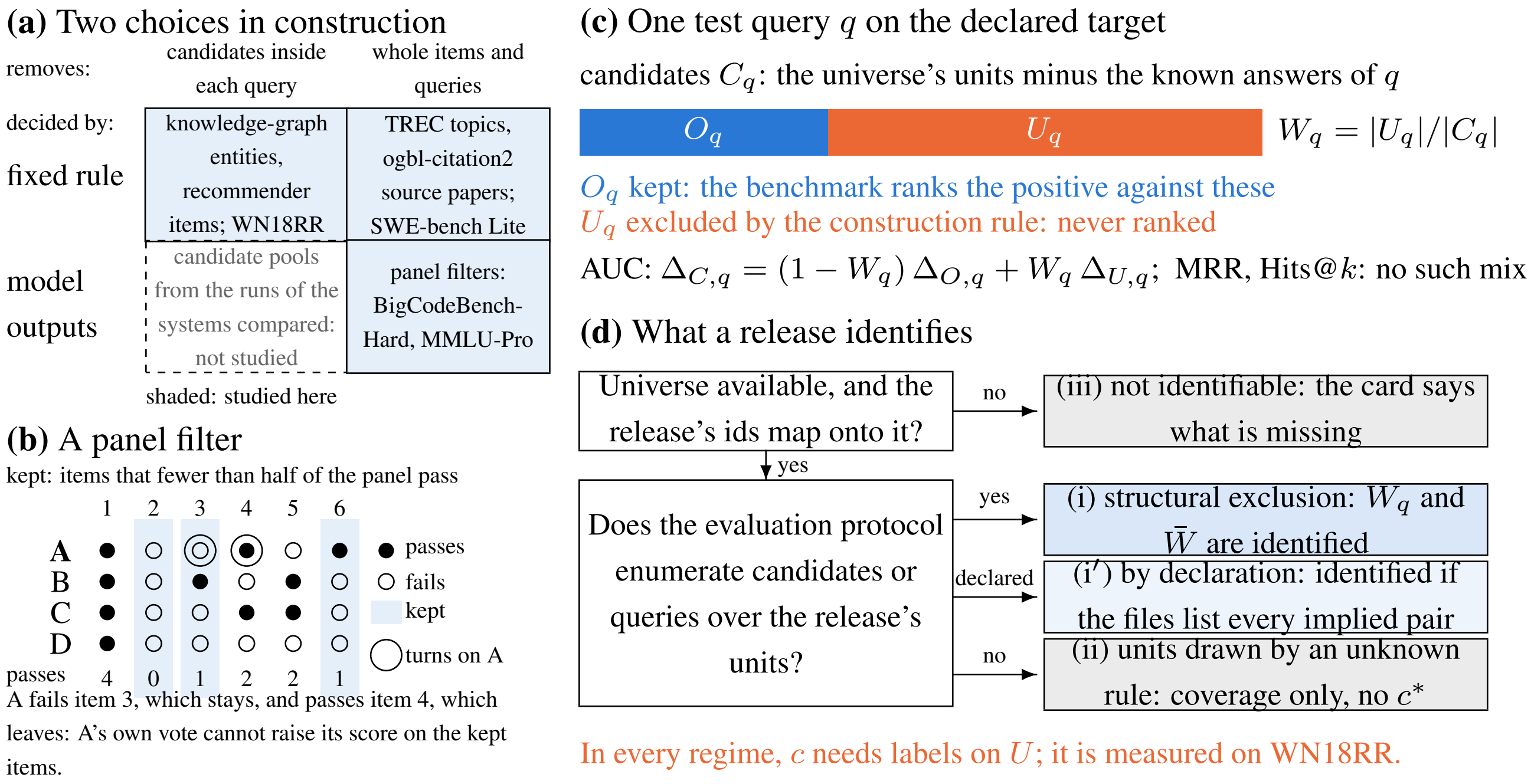
\begin{figure}[tbp]
  \centering
  \resizebox{\textwidth}{!}{\definecolor{bsblue}{HTML}{2A78D6}%
\definecolor{bsorange}{HTML}{EB6834}%
\setlength{\unitlength}{1pt}%
\begin{picture}(382,200)
\put(0,194){\makebox(0,0)[l]{\footnotesize\textbf{(a)}\ Two choices in construction}}
\put(0,182){\makebox(0,0)[l]{\tiny removes:}}
\put(62,182){\makebox(0,0){\parbox{50pt}{\centering\tiny candidates inside each query}}}
\put(114,182){\makebox(0,0){\parbox{50pt}{\centering\tiny whole items and queries}}}
\put(0,168){\makebox(0,0)[l]{\tiny decided by:}}
\put(0,155){\makebox(0,0)[l]{\parbox{34pt}{\raggedright\scriptsize fixed rule}}}
\put(0,121){\makebox(0,0)[l]{\parbox{34pt}{\raggedright\scriptsize model outputs}}}
\put(36,138){\textcolor{bsblue!12}{\rule{52pt}{34pt}}}
\put(36,138){\framebox(52,34){\parbox{48pt}{\centering\tiny knowledge-graph entities, recommender items; WN18RR}}}
\put(88,138){\textcolor{bsblue!12}{\rule{52pt}{34pt}}}
\put(88,138){\framebox(52,34){\parbox{48pt}{\centering\tiny TREC topics, ogbl-citation2 source papers; SWE-bench Lite}}}
\put(36,104){\dashbox{2}(52,34){\parbox{48pt}{\centering\tiny\textcolor{black!60}{candidate pools from the runs of the systems compared: not studied}}}}
\put(88,104){\textcolor{bsblue!12}{\rule{52pt}{34pt}}}
\put(88,104){\framebox(52,34){\parbox{48pt}{\centering\tiny panel filters: BigCodeBench-Hard, MMLU-Pro}}}
\put(36,98){\makebox(0,0)[l]{\tiny shaded: studied here}}
\put(0,86){\makebox(0,0)[l]{\footnotesize\textbf{(b)}\ A panel filter}}
\put(0,77){\makebox(0,0)[l]{\tiny kept: items that fewer than half of the panel pass}}
\put(33,20){\textcolor{bsblue!12}{\rule{10pt}{46pt}}}
\put(45,20){\textcolor{bsblue!12}{\rule{10pt}{46pt}}}
\put(81,20){\textcolor{bsblue!12}{\rule{10pt}{46pt}}}
\put(26,69){\makebox(0,0){\tiny 1}}
\put(38,69){\makebox(0,0){\tiny 2}}
\put(50,69){\makebox(0,0){\tiny 3}}
\put(62,69){\makebox(0,0){\tiny 4}}
\put(74,69){\makebox(0,0){\tiny 5}}
\put(86,69){\makebox(0,0){\tiny 6}}
\put(14,58){\makebox(0,0){\scriptsize \textbf{A}}}
\put(26,58){\circle*{4}}
\put(38,58){\circle{4}}
\put(50,58){\circle{4}}
\put(62,58){\circle*{4}}
\put(74,58){\circle{4}}
\put(86,58){\circle*{4}}
\put(14,50){\makebox(0,0){\scriptsize B}}
\put(26,50){\circle*{4}}
\put(38,50){\circle{4}}
\put(50,50){\circle*{4}}
\put(62,50){\circle{4}}
\put(74,50){\circle*{4}}
\put(86,50){\circle{4}}
\put(14,42){\makebox(0,0){\scriptsize C}}
\put(26,42){\circle*{4}}
\put(38,42){\circle{4}}
\put(50,42){\circle{4}}
\put(62,42){\circle*{4}}
\put(74,42){\circle*{4}}
\put(86,42){\circle{4}}
\put(14,34){\makebox(0,0){\scriptsize D}}
\put(26,34){\circle*{4}}
\put(38,34){\circle{4}}
\put(50,34){\circle{4}}
\put(62,34){\circle{4}}
\put(74,34){\circle{4}}
\put(86,34){\circle{4}}
\put(50,58){\circle{8}}
\put(62,58){\circle{8}}
\put(0,25){\makebox(0,0)[l]{\tiny passes}}
\put(26,25){\makebox(0,0){\tiny 4}}
\put(38,25){\makebox(0,0){\tiny 0}}
\put(50,25){\makebox(0,0){\tiny 1}}
\put(62,25){\makebox(0,0){\tiny 2}}
\put(74,25){\makebox(0,0){\tiny 2}}
\put(86,25){\makebox(0,0){\tiny 1}}
\put(98,58){\circle*{4}}\put(103,58){\makebox(0,0)[l]{\tiny passes}}
\put(98,50){\circle{4}}\put(103,50){\makebox(0,0)[l]{\tiny fails}}
\put(94,39){\textcolor{bsblue!12}{\rule{8pt}{6pt}}}\put(103,42){\makebox(0,0)[l]{\tiny kept}}
\put(98,31){\circle{8}}\put(103,31){\makebox(0,0)[l]{\tiny turns on A}}
\put(0,11){\makebox(0,0)[l]{\parbox{140pt}{\raggedright\tiny A fails item 3, which stays, and passes item 4, which leaves: A\textquoteright s own vote cannot raise its score on the kept items.}}}
\put(148,194){\makebox(0,0)[l]{\footnotesize\textbf{(c)}\ One test query $q$ on the declared target}}
\put(148,180){\makebox(0,0)[l]{\scriptsize candidates $C_q$: the universe's units minus the known answers of $q$}}
\put(148,160){\textcolor{bsblue}{\rule{64pt}{12pt}}}
\put(212,160){\textcolor{bsorange}{\rule{112pt}{12pt}}}
\put(180,166){\makebox(0,0){\scriptsize\textcolor{white}{$O_q$}}}
\put(268,166){\makebox(0,0){\scriptsize\textcolor{white}{$U_q$}}}
\put(328,166){\makebox(0,0)[l]{\scriptsize $W_q=|U_q|/|C_q|$}}
\put(148,151){\makebox(0,0)[l]{\scriptsize\textcolor{bsblue}{$O_q$ kept: the benchmark ranks the positive against these}}}
\put(148,142){\makebox(0,0)[l]{\scriptsize\textcolor{bsorange}{$U_q$ excluded by the construction rule: never ranked}}}
\put(148,130){\makebox(0,0)[l]{\scriptsize AUC: $\Delta_{C,q}=(1-W_q)\,\Delta_{O,q}+W_q\,\Delta_{U,q}$;\enspace MRR, Hits@$k$: no such mix}}
\put(148,114){\makebox(0,0)[l]{\footnotesize\textbf{(d)}\ What a release identifies}}
\put(148,84){\framebox(96,20){\parbox{92pt}{\centering\scriptsize Universe available, and the release's ids map onto it?}}}
\put(244,94){\vector(1,0){22}}
\put(255,97){\makebox(0,0)[b]{\tiny no}}
\put(268,85){\textcolor{black!8}{\rule{114pt}{18pt}}}
\put(268,85){\framebox(114,18){\parbox{110pt}{\centering\scriptsize (iii) not identifiable: the card says what is missing}}}
\put(196,84){\vector(0,-1){8}}
\put(199,79){\makebox(0,0)[l]{\tiny yes}}
\put(148,18){\framebox(96,58){\parbox{92pt}{\centering\scriptsize Does the evaluation protocol enumerate candidates or queries over the release's units?}}}
\put(244,66){\vector(1,0){22}}
\put(255,69){\makebox(0,0)[b]{\tiny yes}}
\put(268,57){\textcolor{bsblue!18}{\rule{114pt}{18pt}}}
\put(268,57){\framebox(114,18){\parbox{110pt}{\centering\scriptsize (i) structural exclusion: $W_q$ and $\bar W$ are identified}}}
\put(244,46){\vector(1,0){22}}
\put(255,49){\makebox(0,0)[b]{\tiny declared}}
\put(268,37){\textcolor{bsblue!8}{\rule{114pt}{18pt}}}
\put(268,37){\framebox(114,18){\parbox{110pt}{\centering\scriptsize (i$'$) by declaration: identified if the files list every implied pair}}}
\put(244,26){\vector(1,0){22}}
\put(255,29){\makebox(0,0)[b]{\tiny no}}
\put(268,17){\textcolor{black!8}{\rule{114pt}{18pt}}}
\put(268,17){\framebox(114,18){\parbox{110pt}{\centering\scriptsize (ii) units drawn by an unknown rule: coverage only, no $c^*$}}}
\put(148,5){\makebox(0,0)[l]{\scriptsize\textcolor{bsorange}{In every regime, $c$ needs labels on $U$; it is measured on WN18RR.}}}
\end{picture}%
 }
  \caption{\textbf{Two choices in construction, and what a release identifies.} \textbf{a}, What a construction step
    removes, candidates inside each query or whole items and queries, and what decides the removal, a fixed rule or
    model outputs. Shaded cells are studied here; candidate pools built from the runs of the systems compared are
    not. A human review, as for SWE-bench Verified, is neither a fixed rule nor model outputs. \textbf{b}, A panel
    filter, as a made-up example: an item is kept when fewer than half of the four members pass it. Items 3 and 4
    turn on member A's vote, because the other three members sit at the threshold: the item that A fails is kept and
    the item that A passes is removed, so A's own vote cannot raise its score on the kept items
    (equation~\eqref{eq:own-vote}). \textbf{c}, A test query on the declared target when exclusion acts within
    queries: the release keeps $O_q$, the construction rule excludes $U_q$, and $W_q$ is the excluded share of the
    query's filtered candidates. Per-query AUC mixes the two regions; reciprocal rank and Hits@$k$ do not.
    \textbf{d}, The questions that decide what a release identifies, answered without labels. A break-even $c^*$ also
    needs a metric that decomposes and a $\Delta_O$ measured on the same queries under the same filter.}
  \label{fig:nmi-setup}
\end{figure}

This section and the next treat fixed rules (Fig.~\ref{fig:nmi-setup}a); the two after them treat filtering by model
outputs, under which the kept population depends on which models sit on the panel.

A declared universe $V$ holds the units a claim is meant to cover, at the unit the benchmark evaluates: entities
in link prediction, user--item pairs in recommendation, topics in retrieval and source papers in citation
prediction. A release keeps $O\subseteq V$, and $U = V\setminus O$ is the excluded part. Exclusion acts at one of
two levels. Where the unit is a query, as with topics in TREC and source papers in ogbl-citation2, it removes whole
queries: the test queries split into kept and excluded ones, and if a share $W$ is excluded,
$\Delta_{\mathrm{target}} = (1-W)\Delta_O + W\Delta_U$ for any metric averaged over queries, with $\Delta_O$,
$\Delta_U$ and $\Delta_{\mathrm{target}}$ two models' mean differences over the kept, the excluded and all queries.
Where the unit is a candidate, as with entities in knowledge graphs and items in recommendation, it removes
candidates inside every query; a recommendation $k$-core does both. A test query $q$ then has a positive $p_q$ and
candidates $C_q$, the units of $V$ it ranks, filtered of other known positives. The benchmark scores only the kept
part $O_q = C_q\cap O$ and never sees $U_q = C_q\cap U$. The excluded share of a query is $W_q = |U_q|/|C_q|$, and
$\bar W$ is its mean over the test queries (Fig.~\ref{fig:nmi-setup}c); $\Delta_O$, $\Delta_U$ and
$\Delta_{\mathrm{target}}$ are now the mean differences in per-query AUC on $O_q$, $U_q$ and $C_q$. The benchmark
reports $\Delta_O$; a claim about the target concerns $\Delta_{\mathrm{target}}$.

When candidates are removed within queries, per-query AUC is a mixture of its values on the two regions, and
averaging over queries gives
\begin{equation}
  \Delta_{\mathrm{target}} = \Delta_O - \bar W\,\bar c - \operatorname{Cov}_q(W_q, c_q),
  \label{eq:nmi-decomp}
\end{equation}
where $c_q = \Delta_{O,q} - \Delta_{U,q}$ is the change of the comparison between the regions and $\bar c$ its
mean (Methods). The target difference therefore has the opposite sign to $\Delta_O$ exactly when
$c\cdot\operatorname{sign}(\Delta_O)$ exceeds the break-even
\begin{equation}
  c^* = |\Delta_O|/\bar W ,
  \label{eq:nmi-breakeven}
\end{equation}
where $c$ is the $W_q$-weighted mean of the $c_q$. These are the per-query forms of the identity and threshold
that \citet{li2026constructed} gave for pooled classification. The threshold needs no additional labels on excluded
candidates, and it is exact for a
$\Delta_O$ measured on the same queries under the same filter; for a difference taken from another query set or
filter it carries no guarantee, and $\Delta_O$ has to be recomputed.

Reciprocal rank and Hits@$k$, by which link prediction and recommendation rank models, do not decompose. Without
ties, both depend only on the number of candidates scored above the positive, which adds up over the two regions,
so neither is a mixture of its regional values. The failure is qualitative: there are models $A$, $A'$ and $B$
whose differences in mean reciprocal rank against $B$ are the same and positive on the kept and on the excluded
candidates, yet $A$ is worse than $B$ on all candidates and $A'$ is better (Methods). Under within-query
exclusion, regional mean differences therefore do not in general determine the target ordering in MRR or
Hits@$k$, and no break-even like $c^*$ exists for them; some sufficient conditions remain, such as
$\Delta_O + \Delta_U > 1$ in MRR (Methods). Target claims below are stated in AUC, with MRR reported alongside.

What a release identifies can be decided without labels and without knowing the rule that built it
(Fig.~\ref{fig:nmi-setup}d). In regime (i), structural exclusion, the evaluation protocol enumerates candidates or
queries over the release's units, the universe is available and the release's ids map onto it; an excluded unit
can then never be evaluated, whatever rule built the release, and every $W_q$ follows from the files and the
protocol. In regime (i$'$), the user declares an enumeration that the files must support. In regime (ii), the
release lists drawn units, such as sampled negatives, but not how they were drawn; absence then does not identify
exclusion, and only coverage is identified. In regime (iii), the universe is unavailable or the ids do not map
onto it. In every regime, $c$ needs the models' scores and labels on the excluded part: whether the excluded part
reverses a comparison is measured, not derived. These statements hold however the split was made; when model
outputs make it, the split itself depends on the panel, and the two sections after the next measure what that adds.

\subsection*{How fixed rules restrict evaluation}

Seven of ten widely reused benchmarks exclude, at the unit they evaluate, between a fifth and nearly all of the
population they are built from, and five releases have undocumented restrictions or differ from their stated rules.
Before any was
inspected, we took in each domain the benchmarks of one widely reused anchor (Extended Data Table~\ref{tab:nmi-ten}): the
knowledge-graph benchmarks compared by \citet{rossi2021knowledge}\citep{toutanova2015observed,dettmers2018convolutional},
the link-prediction pair of the Open Graph Benchmark\citep{hu2020open}, the recommendation sets of NGCF and
LightGCN\citep{wang2019neural,he2020lightgcn}, and the passage tasks of TREC Deep
Learning\citep{craswell2020overview,craswell2021overview} 2019 and 2020. Each was coded from its papers, code and release, and
audited with ScopeCard, the tool described below, against a declared universe at the unit it evaluates
(Methods).

\begin{figure}[tbp]
  \centering
  \includegraphics[width=\textwidth]{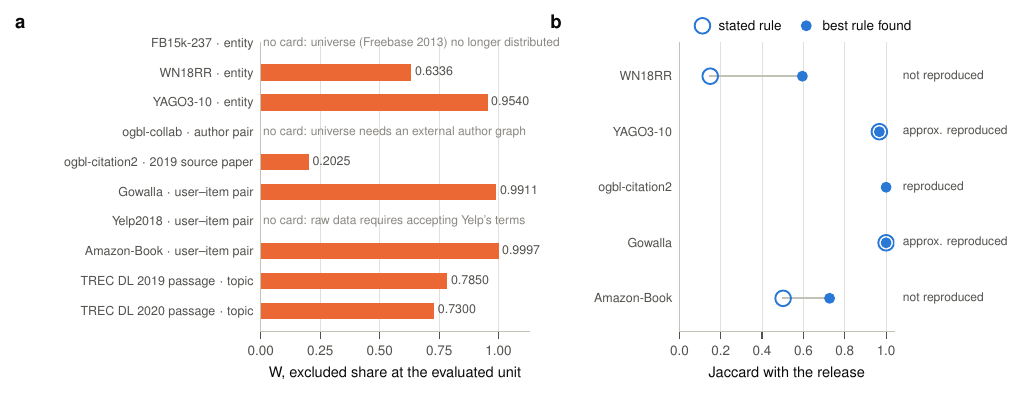}\\[6pt]
  \begin{minipage}[t]{\dimexpr\textwidth/25\relax}\vspace{0pt}\textsf{\textbf{c}}\end{minipage}%
  \begin{minipage}[t]{\dimexpr\textwidth*62/100\relax}\vspace{0pt}\includegraphics[width=\linewidth]{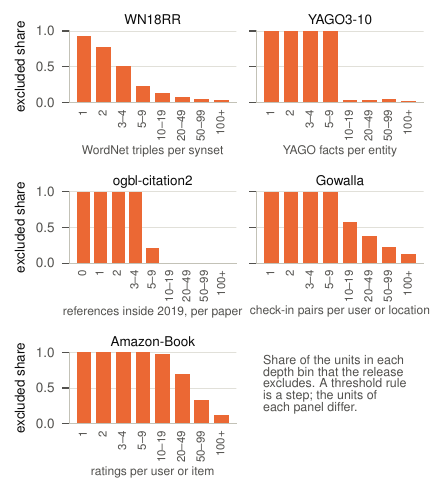}\end{minipage}
  \caption{\textbf{What ten benchmarks exclude.} \textbf{a}, $W$ at the evaluated unit; the three benchmarks in
    regime (iii) are shown in grey with the reason. The unit differs between rows. \textbf{b}, Rule diagnosis: Jaccard similarity of the stated rule (ring) and of the best rule found
    (dot) with the release, with the verdict. \textbf{c}, Excluded share by depth for the five benchmarks whose
    universe has a depth, each in its own unit: a threshold rule is a step, whereas the excluded share of WN18RR's entity
    set, inherited from WN18, falls gradually.}
  \label{fig:nmi-ten}
\end{figure}

The seven fall in regime (i). Their $W$ at the evaluated unit ranges from
\val{0.2025}{card:ogbl-citation2@exclusion.W} for ogbl-citation2's 2019 source papers to
\val{0.9997}{card:amazon-book@exclusion.W_pairs} for Amazon-Book's user--item pairs, and every one except
ogbl-citation2 excludes more than half of its units (Fig.~\ref{fig:nmi-ten}a). $W$ counts a different unit in each
benchmark, so its values compare scope, not quality. Where it is also the share a query sees, no ordering of two
models on the target is identified without assumptions about the excluded part (Methods). The other
\val{3}{src:tool/library/benchmarks.json, why_no_card entries} are in regime (iii): FB15k-237's universe, a 2013
Freebase snapshot, is no longer distributed; ogbl-collab needs an external author graph; and Yelp2018's raw data are
behind Yelp's terms. The construction rule can be decided from published sources for
\val{3}{src:pilot_decidability.tsv, decidability column: verified + decidable, unverified} of the ten.

The five releases are these (rule diagnoses in Fig.~\ref{fig:nmi-ten}b; sources in Methods). WN18's stated
rule\citep{bordes2014semantic} does not reproduce the entity set that WN18RR inherits (Jaccard
\val{0.149}{card:wn18rr@rule_diagnosis.stated.jaccard}; the best simple rule, both ends at depth
\val{5}{card:wn18rr@rule_diagnosis.best.T} or more, reaches \val{0.595}{card:wn18rr@rule_diagnosis.best.jaccard}),
and \val{161}{card:wn18rr@mapping.merged_ids} of its ids each merge synsets of different parts of speech.
FB15k-237 removed \val{71}{txt:results/kg/fb15k237_scope.txt@lost entities that still carry such triples: (\d+)}
entities that still had triples on its kept relations, with
\val{4,555}{txt:results/kg/fb15k237_scope.txt@missing from FB15k-237 train: (\d+)} training triples. YAGO3-10 is
best matched by keeping the facts whose two ends both have at least \val{10}{card:yago3-10@rule_diagnosis.best.T}
facts (Jaccard \val{0.967}{card:yago3-10@rule_diagnosis.best.jaccard}), while its descriptions speak of entities
with at least ten relations\citep{dettmers2018convolutional,akrami2020realistic,rossi2021knowledge}.
ogbl-citation2 evaluates a 2019 paper only if it has at least \val{6}{card:ogbl-citation2@rule_diagnosis.best.T}
references inside the graph (Jaccard \val{1.000}{card:ogbl-citation2@rule_diagnosis.best.jaccard}). Amazon-Book
lies inside the 10-core of its stated source but keeps only part of its users (user Jaccard
\val{0.3318}{pilot:Amazon-Book@verification#2}). Gowalla's stated 10-core is approximately reproduced (Jaccard
\val{0.9999}{card:gowalla@rule_diagnosis.best.jaccard}), and the TREC topic rule, whose bias is
known\citep{craswell2021trec,voorhees2022too}, holds on every judged topic.

The depth profiles of the excluded units differ (Fig.~\ref{fig:nmi-ten}c). YAGO3-10 excludes low-depth entities by
construction, but WN18's excluded share falls gradually with depth, and no threshold searched explains it.
Exclusion is not always a rarity cut, and how hard the excluded part is cannot be read off the rule; it has to be
measured.

WN18RR\citep{dettmers2018convolutional} is a benchmark on which $c$ can be measured, because
WordNet\citep{miller1995wordnet}~\val{3.0}{src:resource version, not a result} holds every synset and triple its
construction left out. The design, estimands and decision rules were prespecified before any model was trained on
the target (Methods). The declared target is every WordNet synset that carries a triple of WN18RR's 11 relations:
\val{112,199}{txt:results/wn18rr_target/summary.txt@universe V \(synsets with >= 1 target triple\): (\d+)}
synsets, of which the \val{41,105}{txt:results/wn18rr_target/summary.txt@\| O: (\d+)} behind WN18RR's entities
form $O$ and the other \val{71,094}{txt:results/wn18rr_target/summary.txt@\| U: (\d+)} form $U$. The excluded
synsets are rarer: their median training degree is
\val{2}{txt:results/wn18rr_target/composition.txt@U: 71094 synsets.*?\n.*?median (\d+)} against
\val{4}{txt:results/wn18rr_target/composition.txt@O: 41105 synsets.*?\n.*?median (\d+)}, and
\val{49.5}{txt:results/wn18rr_target/composition.txt@U: 71094 synsets.*?\n.*?share with degree 1: ([0-9.]+)|pct}\%
of them have degree one, against
\val{7.8}{txt:results/wn18rr_target/composition.txt@O: 41105 synsets.*?\n.*?share with degree 1: ([0-9.]+)|pct}\%.
The models are five embedding models, TransE\citep{bordes2013translating}, ConvE\citep{dettmers2018convolutional},
DistMult\citep{yang2015embedding}, RESCAL\citep{nickel2011three} and ComplEx\citep{trouillon2016complex}, each
with its published configuration\citep{ruffinelli2020you} in LibKGE\citep{broscheit2020libkge} and five seeds; a
popularity baseline, computed from training counts and run once; and NBFNet\citep{zhu2021neural}. Experiment E1 trains and evaluates on WN18RR as published: the MRR seed ranges of
the five embedding models contain their published values, and NBFNet's lies
\val{0.003}{calc:{txt:MEASURED_C_DESIGN.md@official code and config \| ([0-9.]+)}-{max:E1/benchmark/nbfnet/0,1,2,3,4/rr_O}}\textendash{}\val{0.006}{calc:{txt:MEASURED_C_DESIGN.md@official code and config \| ([0-9.]+)}-{min:E1/benchmark/nbfnet/0,1,2,3,4/rr_O}}
below its published value. Experiment E2 trains the other models on the target graph and scores WN18RR's
\val{6,266}{txt:results/wn18rr_target/summary.txt@benchmark test queries \(2 per triple\): (\d+)} test queries
against candidates in $O$ and in $U$, filtered with every target triple. Every comparison below is therefore
between the same E2 models, with the candidates scored as the only difference; Methods reports the effect of
retraining separately. $\Delta_{\mathrm{target}}$ is taken over the fixed benchmark test queries, with
their candidates extended to the target; queries about excluded synsets form a separate, secondary estimand
(Methods). The primary estimand is per-query AUC. Four sensitivity analyses change one
thing each, among them S2, which re-weights $U$ to the training-degree profile of $O$. A difference is resolved when
its \val{95}{src:design constant, MEASURED_C_DESIGN.md Sect. 8}\% triple-bootstrap interval excludes zero, and a
reversal is a pair whose $\Delta_O$ and $\Delta_{\mathrm{target}}$ are resolved with opposite signs.

Admitting the excluded synsets changes every model (Extended Data Fig.~\ref{fig:nmi-models}). On average
\val{0.6337}{v:E2/primary/complex/mean/W} of a query's candidates are excluded synsets. Across the five embedding
models, MRR falls by \val{0.0060}{v:E2/primary/conve/mean/mrr_C_minus_O|neg} (ConvE) to
\val{0.0134}{v:E2/primary/transe/mean/mrr_C_minus_O|neg} (TransE), as adding candidates can only lower it or leave
it unchanged, and for
\val{0.483}{min:E2/primary/transe,conve,distmult,rescal,complex/mean/intrusion}\textendash{}\val{0.534}{max:E2/primary/transe,conve,distmult,rescal,complex/mean/intrusion}
of the queries at least one excluded synset outranks the positive. The sign of AUC on $U$ minus AUC on $O$ depends
on the model, from \val{+0.0211}{v:E2/primary/conve/mean/auc_U_minus_O} for ConvE to
\val{-0.0180}{v:E2/primary/rescal/mean/auc_U_minus_O} for RESCAL; the popularity baseline lies outside all three
ranges (Extended Data Table~\ref{ed:models}). On the degree-matched target the difference is positive for all
\val{6}{count(v>0):E2/S2/transe,conve,distmult,rescal,complex,popularity/mean/auc_U_minus_O}, the baseline included.

\begin{figure}[tbp]
  \centering
  \includegraphics[width=\textwidth]{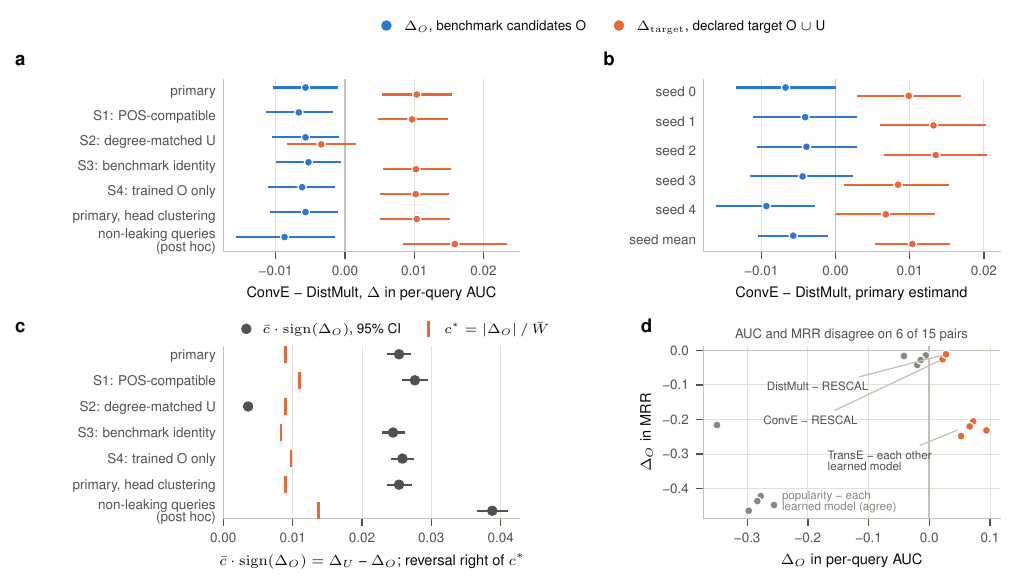}
  \caption{\textbf{ConvE $-$ DistMult.} \textbf{a}, $\Delta_O$ and $\Delta_{\mathrm{target}}$ under the primary
    estimand, the sensitivity analyses, head-synset clustering and the post hoc non-leaking queries. \textbf{b},
    Per seed. \textbf{c}, $\bar c\cdot\operatorname{sign}(\Delta_O) = \Delta_U - \Delta_O$ against $c^*$. The exact
    criterion, equation~\eqref{eq:nmi-breakeven}, uses the $W_q$-weighted $c$, which differs from $\bar c$ by
    $\val{1.2}{calc:1000000*{v:E2/primary/conve-distmult/mean/approx_err|neg}/{v:E2/primary/conve-distmult/mean/W}}\times10^{-6}$
    under the primary estimand and by at most
    $\val{7.2}{calc:10000*{v:E2/S1/conve-distmult/mean/approx_err|neg}/{v:E2/S1/conve-distmult/mean/W}}\times10^{-4}$
    in the other rows (S1); in every row both lie on the same side of $c^*$. To the right of $c^*$, the point
    estimates of $\Delta_O$ and $\Delta_{\mathrm{target}}$ have opposite signs; a reversal also needs both to be
    resolved (\textbf{a}). \textbf{d}, The AUC difference against the MRR difference on $O$ for all 15
    pairs; the pairs on which the two metrics disagree are in orange.}
  \label{fig:nmi-pair}
\end{figure}

Under the prespecified criterion, with unadjusted per-pair intervals, one of the
\val{15}{count(excl0):E2/primary/*-*/mean/delta_O} pairs reverses (Fig.~\ref{fig:nmi-pair}). For ConvE
$-$ DistMult, $\Delta_O = \val{-0.0057}{v:E2/primary/conve-distmult/mean/delta_O}$
$[\val{-0.0103}{v:E2/primary/conve-distmult/mean/delta_O.lo},\val{-0.0011}{v:E2/primary/conve-distmult/mean/delta_O.hi}]$
and $\Delta_{\mathrm{target}} = \val{+0.0104}{v:E2/primary/conve-distmult/mean/delta_target}$
$[\val{+0.0055}{v:E2/primary/conve-distmult/mean/delta_target.lo},\val{+0.0153}{v:E2/primary/conve-distmult/mean/delta_target.hi}]$:
DistMult leads on the benchmark and ConvE on the target. ConvE's advantage over DistMult on the excluded synsets is
large enough to reverse the order: from the kept to the excluded synsets the difference moves by
$\Delta_U - \Delta_O = \val{+0.0253}{v:E2/primary/conve-distmult/mean/delta_U_minus_O}$, more than the break-even
$c^* = \val{0.0090}{v:E2/primary/conve-distmult/mean/c_star}$ of equation~\eqref{eq:nmi-breakeven}. The threshold
needs no additional labels on the excluded
synsets, and its $\bar W$ is the value that ScopeCard reports for WN18RR (Extended Data
Table~\ref{ed:validation}); measuring $\Delta_U - \Delta_O$ needed models trained on the target. Both differences
stay resolved with opposite signs when candidates are restricted to the part of speech that fits the relation (S1), when synsets are keyed as WN18RR
keys them (S3), when $O$ keeps only trained synsets (S4), under clustering by head synset, and on the queries
whose test triple's reverse is not an E2 training triple (post hoc). $\Delta_{\mathrm{target}}$ is resolved and positive in all
\val{5}{count(lo>0):E2/primary/conve-distmult/0,1,2,3,4/delta_target} seeds, $\Delta_O$ is resolved in
\val{2}{count(excl0):E2/primary/conve-distmult/0,1,2,3,4/delta_O} of them. The reversal does not survive degree
reweighting: under S2, $\Delta_{\mathrm{target}} = \val{-0.0034}{v:E2/S2/conve-distmult/mean/delta_target}$
$[\val{-0.0082}{v:E2/S2/conve-distmult/mean/delta_target.lo},\val{+0.0014}{v:E2/S2/conve-distmult/mean/delta_target.hi}]$,
so the comparison is sensitive to the degree composition of the target.

The metric matters too. In MRR the pair does not reverse: DistMult leads on $O$
(\val{-0.0146}{v:E2/primary/conve-distmult/mean/delta_mrr_O}) and on all candidates
(\val{-0.0087}{v:E2/primary/conve-distmult/mean/delta_mrr_C}), both resolved. On $O$ itself, AUC and MRR order
\val{6}{n:E2/primary/transe-conve,transe-distmult,transe-rescal,transe-complex,conve-rescal,distmult-rescal/mean/delta_mrr_O}
of the 15 pairs oppositely; TransE has the highest AUC on $O$,
\val{0.970}{max:E2/primary/transe,conve,distmult,rescal,complex/mean/auc_O}, and the lowest MRR among the learned
models, \val{0.238}{min:E2/primary/transe,conve,distmult,rescal,complex/mean/rr_O}. Which model is better on
WN18RR's population therefore depends on whether that population is the benchmark, the full target or the
degree-matched target, and on whether candidates are scored by AUC or by MRR.

Two further results bound these findings. NBFNet, trained on WN18RR and deployed on the target graph without
retraining, loses the most MRR when the excluded synsets are admitted,
\val{-0.0205}{v:deploy/primary/nbfnet/mean/mrr_C_minus_O}
$[\val{-0.0221}{v:deploy/primary/nbfnet/mean/mrr_C_minus_O.lo}$, $\val{-0.0189}{v:deploy/primary/nbfnet/mean/mrr_C_minus_O.hi}]$,
but its AUC on $U$ minus AUC on $O$ ranges from \val{-0.0027}{min:deploy/primary/nbfnet/0,1,2,3,4/auc_U_minus_O} to
\val{+0.0047}{max:deploy/primary/nbfnet/0,1,2,3,4/auc_U_minus_O} across seeds. And
\val{2,172}{v:E2/posthoc_bench_test:leak/complex/0/auc_C.n} of the
\val{6,266}{v:E2/posthoc_bench_test:all/complex/0/auc_C.n} test queries
(\val{35}{calc:100*{v:E2/posthoc_bench_test:leak/complex/0/auc_C.n}/{v:E2/posthoc_bench_test:all/complex/0/auc_C.n}}\%)
come from test triples of WN18RR's symmetric relations whose reverse is an E2 training triple; the reversal holds
without them.

\subsection*{Filtering by model outputs: mechanism and reach}

When the outputs of a panel of models decide which items stay, the kept set depends on who sits on the panel.
Released benchmarks use threshold rules of this kind: among its criteria, BigCodeBench-Hard keeps tasks that fewer
than half of a panel of models solve\citep{zhuo2025bigcodebench}, and MMLU-Pro removed questions that more than half of
eight models answered correctly\citep{wang2024mmlupro}. Whether a member $m$ votes, rather than a model $u$ outside the
panel, changes the kept set only on items where the rest of the panel sits at the threshold and $m$ and $u$ disagree:
with $m$ on the panel, the items that $m$ fails are kept and the items that $m$ solves are removed
(Fig.~\ref{fig:nmi-setup}b). When the outputs
that vote are also the outputs that are scored, it follows that $m$'s own vote cannot raise its score on the kept
set, whatever the panel, the threshold and the item pool (Methods, equation~\eqref{eq:own-vote}). The sign of this
own-vote penalty is fixed by algebra; only its size is empirical. For outputs that did not vote, such as $m$'s
answers in another prompt format or the answers of a related model, the same identity holds with their mean outcomes
on the items that enter and leave, and the sign is no longer fixed: the change is positive when these outputs tend to
fail where $m$ fails and pass where $m$ passes.

\begin{figure}[tbp]
  \centering
  \includegraphics[width=\textwidth]{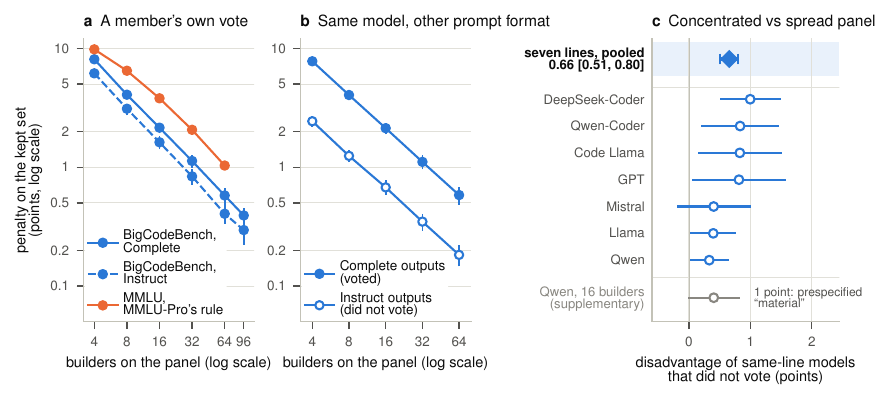}
  \caption{\textbf{Filtering by model outputs on archived runs.} \textbf{a}, The own-vote penalty, a member's score on
    the kept set when it is replaced by an outside model minus its score when another member is replaced instead,
    against the number of builders on random panels under the published rules. Both axes are logarithmic; the lines
    join the measured panel sizes. \textbf{b}, Panels vote with BigCodeBench's Complete outputs; members are scored on
    the Complete outputs that voted (filled) and on their own Instruct outputs, which did not vote (open). Both series
    are the penalty itself; the excess penalty in the text also subtracts the change of ten models that voted in neither
    panel. \textbf{c}, How much less the models of a line that never voted score on the kept set of a concentrated
    panel (eight builders, four of them from the line) than on that of a spread panel (one of the four kept, the other
    three replaced by builders of lines not yet on the panel), net of the same difference for models of unchanged
    lines with similar Instruct pass rates; votes from Complete outputs, scores from Instruct outputs (Methods). The
    diamond, with its value under its label, is the pooled estimate over the seven lines and its
    \val{95}{src:design constant, interval level}\% percentile interval; each circle is one line with its
    simultaneous \val{95}{src:design constant, interval level}\% interval over the eight line contrasts. Qwen at 16
    builders (grey) is a supplementary contrast. Dashed: one point, which the prespecified reading calls material. In
    \textbf{a} and \textbf{b}, bars are pointwise \val{95}{src:design constant, interval level}\% percentile intervals
    from the bootstraps over tasks or questions (Methods). Every value drawn is listed with its record in the
    reproduction package (\texttt{fig\_filters\_values.tsv}).}
  \label{fig:filters}
\end{figure}

On archived item-level runs, with random panels under the published rules, the own-vote penalty fell as panels grew
(Fig.~\ref{fig:filters}a).
On BigCodeBench's Complete prompts a member lost \val{4.07}{txt:results/panel_filter/summary_panel_size.md@\n.. Complete[\s\S]*?\n\| 8 \| [^|]*\| [^|]*\| \+([\d.]+)} points at 8 builders
(\val{95}{src:design constant, interval level}\% interval \val{3.71}{txt:results/panel_filter/summary_panel_size.md@\n.. Complete[\s\S]*?\n\| 8 \| [^|]*\| [^|]*\| \+[\d.]+ \[\+([\d.]+),}\textendash{}\val{4.37}{txt:results/panel_filter/summary_panel_size.md@\n.. Complete[\s\S]*?\n\| 8 \| [^|]*\| [^|]*\| \+[\d.]+ \[\+[\d.]+, \+([\d.]+)\]}) and
\val{0.39}{txt:results/panel_filter/summary_panel_size.md@\n.. Complete[\s\S]*?\n\| 96 \| [^|]*\| [^|]*\| \+([\d.]+)} at 96; on MMLU under MMLU-Pro's rule, \val{6.48}{txt:results/mmlu_reconstruction/summary_panel_size.md@\n\| 8 \| [^|]*\| [^|]*\| \+([\d.]+)} at 8 and \val{1.03}{txt:results/mmlu_reconstruction/summary_panel_size.md@\n\| 64 \| [^|]*\| [^|]*\| \+([\d.]+)} at
64. The product of penalty and panel size stayed between \val{32.3}{txt:results/panel_filter/summary_panel_size.md@\n.. Complete[\s\S]*?Own participation times the number of builders[^\n]*p = 4: (\d+\.\d+)} and
\val{37.7}{txt:results/panel_filter/summary_panel_size.md@\n.. Complete[\s\S]*?Own participation times the number of builders[^\n]*p = 96: (\d+\.\d+)} on BigCodeBench's Complete prompts and between \val{24.6}{txt:results/panel_filter/summary_panel_size.md@\n.. Instruct[\s\S]*?Own participation times the number of builders[^\n]*p = 4: (\d+\.\d+)} and
\val{28.4}{txt:results/panel_filter/summary_panel_size.md@\n.. Instruct[\s\S]*?Own participation times the number of builders[^\n]*p = 96: (\d+\.\d+)} on its Instruct prompts, from 4 to 96 builders, but on MMLU it rose from
\val{39.2}{txt:results/mmlu_reconstruction/summary_panel_size.md@Own participation times the number of builders[^\n]*p = 4: (\d+\.\d+)} at 4 builders to \val{66.0}{txt:results/mmlu_reconstruction/summary_panel_size.md@Own participation times the number of builders[^\n]*p = 32: (\d+\.\d+)} at 32 and 64. When the other builders of a member's line
were counted with it, the MMLU penalty fell more slowly still (penalty times panel size \val{40}{calc:4*{txt:results/mmlu_reconstruction/summary_panel_size.md@\n\| 4 \| [^|]*\| [^|]*\| [^|]*\| [^|]*\| \+([\d.]+)}} at 4
builders and \val{107}{calc:64*{txt:results/mmlu_reconstruction/summary_panel_size.md@\n\| 64 \| [^|]*\| [^|]*\| [^|]*\| [^|]*\| \+([\d.]+)}} at 64), because larger random panels held more models of one line. The
penalty thus fell at different rates on the two benchmarks and from different levels in the two prompt formats, and
the part added by the other builders of a member's line did not fall with the member's own: across panel sizes it
stayed between
\val{0.09}{txt:results/panel_filter/summary_panel_size.md@\n.. Complete[\s\S]*?\n\| 4 \| [^|]*\| [^|]*\| [^|]*\| [^|]*\| [^|]*\| \+([\d.]+)} and
\val{0.25}{txt:results/panel_filter/summary_panel_size.md@\n.. Complete[\s\S]*?\n\| 96 \| [^|]*\| [^|]*\| [^|]*\| [^|]*\| [^|]*\| \+([\d.]+)} points on BigCodeBench's Complete prompts and between
\val{0.25}{txt:results/mmlu_reconstruction/summary_panel_size.md@\n\| 4 \| [^|]*\| [^|]*\| [^|]*\| [^|]*\| [^|]*\| \+([\d.]+)} and
\val{1.26}{txt:results/mmlu_reconstruction/summary_panel_size.md@\n\| 32 \| [^|]*\| [^|]*\| [^|]*\| [^|]*\| [^|]*\| \+([\d.]+)} points on MMLU.

The penalty reached outputs that never voted (Fig.~\ref{fig:filters}b). With panels voting on BigCodeBench's Complete outputs and members
scored on their Instruct outputs, which took no part in the selection, members lost
\val{1.24}{txt:results/separated_scoring/effects.tsv@complete_to_instruct\t8\ttarget\t(\S+)|pct} points at 8 builders, against
\val{4.05}{txt:results/separated_scoring/effects.tsv@complete_to_instruct\t8\tsource\t(\S+)|pct} on the outputs that voted; their excess over models
outside the panel was \val{1.22}{txt:results/separated_scoring/effects.tsv@complete_to_instruct\t8\texcess\t(\S+)|pct} points (simultaneous \val{95}{src:design constant, interval level}\%
interval \val{1.04}{txt:results/separated_scoring/effects.tsv@complete_to_instruct\t8\texcess\t\S*\t\S*\t\S*\t(\S+)|pct}\textendash{}\val{1.41}{txt:results/separated_scoring/effects.tsv@complete_to_instruct\t8\texcess\t\S*\t\S*\t\S*\t\S*\t(\S+)|pct}).
Over panel sizes from 4 to 64 and in both directions, the penalty on the other format was
\val{31}{calc:100*{txt:results/separated_scoring/effects.tsv@complete_to_instruct\t8\ttarget\t(\S+)}/{txt:results/separated_scoring/effects.tsv@complete_to_instruct\t8\tsource\t(\S+)}}\textendash{}\val{34}{calc:100*{txt:results/separated_scoring/effects.tsv@instruct_to_complete\t4\ttarget\t(\S+)}/{txt:results/separated_scoring/effects.tsv@instruct_to_complete\t4\tsource\t(\S+)}}\% of the penalty on the
outputs that voted.

The reach extends to models that did not vote (Fig.~\ref{fig:filters}c). On BigCodeBench, at a fixed panel size, we
compared for each of seven model lines an eight-model panel with four members from the line against a panel that kept
one of them and replaced the other three, one by one, by the models of lines not yet on the panel nearest in Complete
pass rate, so that the two panels' mean pass rates matched approximately; votes came from Complete
outputs and scores from Instruct outputs (Methods). The line's models that never voted were at
a disadvantage of \val{0.66}{txt:results/panel_composition/records.tsv@primary\tT8\testimate\t(\S+)|pct} points (\val{95}{src:design constant, interval level}\% interval
\val{0.51}{txt:results/panel_composition/records.tsv@primary\tT8\tlo\t(\S+)|pct}\textendash{}\val{0.80}{txt:results/panel_composition/records.tsv@primary\tT8\thi\t(\S+)|pct}) on the concentrated panel's kept set,
relative to models of unaffected lines with similar Instruct pass rates: detectable but small by the prespecified reading. Under
simultaneous intervals the disadvantage was detected for six of the seven lines, at
\val{0.33}{txt:results/panel_composition/records.tsv@line_8\tqwen\testimate\t(\S+)|pct} to \val{0.997}{txt:results/panel_composition/records.tsv@line_8\tdeepseek-coder\testimate\t(\S+)|pct} points;
the exception was Mistral
(\val{0.40}{txt:results/panel_composition/records.tsv@line_8\tmistral\testimate\t(\S+)|pct}). For Qwen, the one line large enough for a supplementary 16-model
contrast, it was detected with eight models but not with sixteen
(\val{0.40}{txt:results/panel_composition/records.tsv@line_16\tqwen\testimate\t(\S+)|pct}). On the spread panel, the evaluation models of the lines that gained a
seat, pooled, changed by \val{-0.05}{txt:results/panel_composition/records.tsv@line_8\tcodellama\tcost_of_spreading\t(\S+)|pct} to
\val{-0.26}{txt:results/panel_composition/records.tsv@line_8\tmistral\tcost_of_spreading\t(\S+)|pct} points relative to their controls for six focal lines,
and by \val{+0.08}{txt:results/panel_composition/records.tsv@line_8\tgpt\tcost_of_spreading\t(\S+)|pct} when GPT was the focal line. The two kept sets differed in no consistent direction: by \val{-6.3}{calc:1*({txt:results/panel_composition/records.tsv@kept_8_spread\tcodellama\tkept\t(\S+)}-{txt:results/panel_composition/records.tsv@kept_8_conc\tcodellama\tkept\t(\S+)})} to \val{+5.3}{calc:1*({txt:results/panel_composition/records.tsv@kept_16_spread\tqwen\tkept\t(\S+)}-{txt:results/panel_composition/records.tsv@kept_16_conc\tqwen\tkept\t(\S+)})}
tasks in size, \val{-0.58}{calc:100*({txt:results/panel_composition/records.tsv@kept_8_spread\tcodellama\tdifficulty\t(\S+)}-{txt:results/panel_composition/records.tsv@kept_8_conc\tcodellama\tdifficulty\t(\S+)})} to
\val{+0.38}{calc:100*({txt:results/panel_composition/records.tsv@kept_16_spread\tqwen\tdifficulty\t(\S+)}-{txt:results/panel_composition/records.tsv@kept_16_conc\tqwen\tdifficulty\t(\S+)})} points in mean pass rate and
\val{-0.0125}{calc:1*({txt:results/panel_composition/records.tsv@kept_8_spread\tcodellama\tdiscrimination\t(\S+)}-{txt:results/panel_composition/records.tsv@kept_8_conc\tcodellama\tdiscrimination\t(\S+)})} to \val{+0.0080}{calc:1*({txt:results/panel_composition/records.tsv@kept_16_spread\tqwen\tdiscrimination\t(\S+)}-{txt:results/panel_composition/records.tsv@kept_16_conc\tqwen\tdiscrimination\t(\S+)})}
in item discrimination.

\subsection*{The order on released filtered subsets, and its boundaries}

On three released subsets of coding benchmarks, built by a scope rule (SWE-bench
Lite\citep{jimenez2024swebench,jimenez2024swebenchlite}), a human validity review (SWE-bench
Verified\citep{chowdhury2024swebenchverified}) and a difficulty rule over model results
(BigCodeBench-Hard\citep{zhuo2025bigcodebench}), levels and gaps changed between the kept part and its parent, but no
reversal was established under simultaneous intervals, and the number of pairs with opposite signs stayed within what
random subsets of the same size give (Table~\ref{tab:order}).

MMLU-Pro's release identifies which questions of the original MMLU\citep{hendrycks2021measuring} it retained after its model
filter and its later review; matched by text, they form the kept set below. The panel's votes are not available,
and the release does not tell apart the questions that the filter removed and those that the review removed, so the
kept set reflects both. Five of the eight filter models have archived item-level runs on
MMLU\citep{beeching2023open}. Scored with these runs, which are not the runs that voted, the five
filter models did \val{1.68}{txt:results/mmlu_reconstruction/summary.md@Penalty P = \+([\d.]+) points} points worse on the kept questions, relative to
their scores on all questions, than systems of similar overall accuracy that took no part (\val{95}{src:design constant, interval level}\% interval
\val{1.39}{txt:results/mmlu_reconstruction/summary.md@Penalty P = \+[\d.]+ points\*\* \(question bootstrap 95. interval \+([\d.]+) to}\textendash{}\val{1.97}{txt:results/mmlu_reconstruction/summary.md@Penalty P = \+[\d.]+ points\*\* \(question bootstrap 95. interval \+[\d.]+ to \+([\d.]+)\)};
none of \val{53,130}{txt:results/mmlu_reconstruction/summary.md@reference tail share: 0 of (\d+) subsets} same-size subsets of a placebo pool reached
it). The penalty differed between models: it was largest for
Llama-2-13b (\val{+5.36}{txt:results/mmlu_reconstruction/summary.md@meta-llama__Llama-2-13b-hf \+([\d.]+) \[}) and Llama-2-7b
(\val{+2.02}{txt:results/mmlu_reconstruction/summary.md@meta-llama__Llama-2-7b-hf \+([\d.]+) \[}), about one point for Mistral-7B and Gemma-7B, and
negative for Yi-6B (\val{-1.06}{txt:results/mmlu_reconstruction/summary.md@01-ai__Yi-6B \+?(-?[\d.]+) \[}).
On MMLU-Pro's own questions (with up to ten options), where six of the filter models have outputs and the controls were chosen
more widely (Methods), an estimate of the same form was
\val{1.57}{txt:results/mmlupro_filter/summary.md@Penalty P = \+([\d.]+) points} points, with a question-bootstrap
\val{95}{src:design constant, interval level}\% interval of
\val{0.63}{txt:results/mmlupro_filter/summary.md@Penalty P = \+[\d.]+ points\*\* \(question bootstrap 95. interval \+([\d.]+) to}\textendash{}\val{2.60}{txt:results/mmlupro_filter/summary.md@Penalty P = \+[\d.]+ points\*\* \(question bootstrap 95. interval \+[\d.]+ to \+([\d.]+)\)}
that excludes zero; \val{687}{txt:results/mmlupro_filter/summary.md@reference tail share: (\d+) of 5005} of \val{5,005}{txt:results/mmlupro_filter/summary.md@reference tail share: 687 of (\d+)}
same-size subsets of the placebo pool reached it, so the value is not unusual among the reference systems.

On the kept MMLU questions the order changed more than random subsets give, but mostly among systems that score
near chance on them. Among the five filter models and 70 clean systems, \val{261}{txt:results/mmlu_reconstruction/summary_pairs.md@Kendall tau_b between the scores on kept and on all questions [\d.]+; (\d+) of} of \val{2,775}{txt:results/mmlu_reconstruction/summary_pairs.md@- 75 systems, (\d+) pairs, \d+ tied} pairs had opposite signs on the
kept and on all questions, against \val{16}{txt:results/mmlu_reconstruction/summary_pairs.md@\(261 against (\d+) to}\textendash{}\val{39}{txt:results/mmlu_reconstruction/summary_pairs.md@\(261 against \d+ to (\d+)\)} for random subsets with the same number of questions per subject,
and no reversal was established under simultaneous intervals. The kept set leaves the weaker part of the field near
guessing: the \val{31}{calc:{txt:results/mmlu_reconstruction/summary_pairs_posthoc.md@\| below 30. \| (\d+) \|}+{txt:results/mmlu_reconstruction/summary_pairs_posthoc.md@\| 30. to 40. \| (\d+) \|}} clean systems below \val{40}{src:bin edge of the floor description, summary_pairs_posthoc.md Section 1}\% on all questions scored \val{24.2}{txt:results/mmlu_reconstruction/summary_pairs_posthoc.md@\| below 30. \| \d+ \| ([\d.]+). to}\textendash{}\val{27.3}{txt:results/mmlu_reconstruction/summary_pairs_posthoc.md@\| 30. to 40. \| \d+ \| [\d.]+. to ([\d.]+).}\% on the kept ones,
against \val{25}{src:chance level with four answer options}\% for guessing. Among the \val{35}{txt:results/mmlu_reconstruction/summary_pairs_posthoc.md@\| A \| 30. \| (\d+) \|} systems at or above \val{30}{src:floor cut chosen after the run, MMLU_RECONSTRUCTION_DESIGN.md Section 11 (post hoc)}\% on the kept questions (a cut chosen
after the run), \val{7}{txt:results/mmlu_reconstruction/summary_pairs_posthoc.md@\| A \| 30. \| \d+ \| \d+ \| (\d+) \|} of \val{595}{txt:results/mmlu_reconstruction/summary_pairs_posthoc.md@\| A \| 30. \| \d+ \| (\d+) \|} pairs changed sign, within the reference of \val{1}{txt:results/mmlu_reconstruction/summary_pairs_posthoc.md@\| A \| 30. \| \d+ \| \d+ \| \d+ \| (\d+) to}\textendash{}\val{8}{txt:results/mmlu_reconstruction/summary_pairs_posthoc.md@\| A \| 30. \| \d+ \| \d+ \| \d+ \| \d+ to (\d+) \|}. With the
\val{79}{calc:{txt:results/mmlu_reconstruction/summary_pairs.md@Population K: the systems of known provenance[^\n]*\n- (\d+) systems}-{txt:results/mmlu_reconstruction/summary_pairs.md@Population A: the filter models and the clean pool[^\n]*\n- (\d+) systems}} systems descended from the filter models added, \val{36}{txt:results/mmlu_reconstruction/summary_pairs.md@\*\*reversals (\d+)\*\*; newly resolved 1194} reversals were established, and in each of them
the system that fell behind descends from Llama-2-7b or Llama-2-13b, the line that held four of the eight votes.

Because the kept set is harder than random subsets, part of this excess could come from difficulty alone. We therefore
also compared it with random subsets matched on subject and on difficulty estimated from systems outside the scored
set (cross-fitted between halves of the clean pool, eight strata; R1), and in a second reference also on the answer
key (R2; Methods). This analysis is exploratory (Methods). For pairs of clean systems, the kept set gave
\val{113}{txt:results/mmlu_reconstruction/summary_difficulty_reference.md@\| clean.clean, same half \| \d+ \| (\d+) \| [\d.]+ to [\d.]+ \([\d.]+\)} sign changes against \val{46}{txt:results/mmlu_reconstruction/summary_difficulty_reference.md@\| clean.clean, same half \| \d+ \| \d+ \| [\d.]+ to [\d.]+ \([\d.]+\) \| [\d.]+ to [\d.]+ \([\d.]+\) \| ([\d.]+) to [\d.]+ \(}\textendash{}\val{63}{txt:results/mmlu_reconstruction/summary_difficulty_reference.md@\| clean.clean, same half \| \d+ \| \d+ \| [\d.]+ to [\d.]+ \([\d.]+\) \| [\d.]+ to [\d.]+ \([\d.]+\) \| [\d.]+ to ([\d.]+) \(}
in the reference matched on all three features, and all four types of pairs read were above all three references
(Table~\ref{tab:order}). The excess lay among systems near chance on the kept questions and in the panel's line: pairs of two
clean systems at or above the floor gave \val{3}{txt:results/mmlu_reconstruction/summary_difficulty_reference.md@\| clean.clean, same half \| 0\.30 \| both above \| \d+ \| (\d+) \|} against \val{1}{txt:results/mmlu_reconstruction/summary_difficulty_reference.md@\| clean.clean, same half \| 0\.30 \| both above \| \d+ \| \d+ \| [^|]*\| [^|]*\| (\d+) to}\textendash{}\val{5}{txt:results/mmlu_reconstruction/summary_difficulty_reference.md@\| clean.clean, same half \| 0\.30 \| both above \| \d+ \| \d+ \| [^|]*\| [^|]*\| \d+ to (\d+) },
within the matched reference, while pairs of a Llama-2-7b descendant and a clean system gave
\val{315}{txt:results/mmlu_reconstruction/summary_difficulty_reference.md@\| Llama-2-7b \| \d+ \| (\d+) \| [\d.]+ to [\d.]+ \([\d.]+\)} against \val{34}{txt:results/mmlu_reconstruction/summary_difficulty_reference.md@\| Llama-2-7b \| \d+ \| \d+ \| [\d.]+ to [\d.]+ \([\d.]+\) \| [\d.]+ to [\d.]+ \([\d.]+\) \| ([\d.]+) to [\d.]+ \(}\textendash{}\val{44}{txt:results/mmlu_reconstruction/summary_difficulty_reference.md@\| Llama-2-7b \| \d+ \| \d+ \| [\d.]+ to [\d.]+ \([\d.]+\) \| [\d.]+ to [\d.]+ \([\d.]+\) \| [\d.]+ to ([\d.]+) \(}. A
residual difference therefore remains after matching on subject, estimated difficulty and answer key, concentrated at
the guessing floor and in the Llama-2 line. The matched subsets stayed slightly easier than the kept set, and
MMLU-Pro's later review cannot be separated from the panel's choice, so the residual does not identify an effect of the
panel (Methods).

\begin{table}[tbp]
  \caption{\textbf{Order on kept items against reference subsets.} Opposite signs: pairs of systems whose difference
    has opposite signs on the kept items and on all items (for the released subsets, on the kept part and on its
    parent). The parents are SWE-bench's
    \val{2,294}{txt:results/subsets/summary.md@\n.. Case A[\s\S]*?- \d+ systems, (\d+) instances; kept} test
    instances (Lite keeps \val{300}{txt:results/subsets/summary.md@\n.. Case A[\s\S]*?- \d+ systems, \d+ instances; kept (\d+),}),
    the \val{1,699}{txt:results/subsets/summary.md@\n.. Case B[\s\S]*?- \d+ systems, (\d+) instances; kept}
    instances with human annotations (Verified keeps
    \val{500}{txt:results/subsets/summary.md@\n.. Case B[\s\S]*?- \d+ systems, \d+ instances; kept (\d+),}; the other
    \val{595}{txt:results/subsets/summary.md@Test instances without annotations \(outside this case\): (\d+)} test
    instances are not compared) and BigCodeBench's
    \val{1,140}{txt:results/subsets/summary.md@\n.. Case C[\s\S]*?- \d+ systems, (\d+) instances; kept} tasks (Hard
    keeps \val{148}{txt:results/subsets/summary.md@\n.. Case C[\s\S]*?- \d+ systems, \d+ instances; kept (\d+),}). For
    MMLU, all questions are the
    \val{13,147}{txt:results/mmlu_reconstruction/summary_pairs.md@\d+ subjects with (\d+) original questions}
    questions of the
    \val{56}{txt:results/mmlu_reconstruction/summary_pairs.md@; (\d+) subjects with \d+ original questions} subjects
    that have kept questions
    (\val{6,580}{txt:results/mmlu_reconstruction/summary_pairs.md@\d+ subjects with \d+ original questions, of which (\d+) kept}
    of them kept), scored with the kept questions' subject mix. Pairs: all pairs of the systems read; a pair tied on either set does not count as having opposite
    signs. References: the central \val{95}{src:design constant, interval level}\% of the same count over random subsets of the same size (uniform for the
    released subsets; with the kept set's number of questions per subject for MMLU, R0), and over subsets also
    matched on estimated difficulty (R1) and on the answer key (R2). Reversals: pairs resolved with opposite signs under
    simultaneous \val{95}{src:design constant, interval level}\% intervals. Clean pairs in the matched rows lie within one half of the clean pool, whose
    difficulty is estimated from the other half; a pair scored in both folds counts with the mean of its two
    indicators, so a bound of a matched reference can be a half-integer. The floor cut and the matched references are
    exploratory. BigCodeBench-Hard is read in its Complete prompt format. Blank cells: not computed.}
  \label{tab:order}
  \centering\footnotesize
  \begin{tabular}{@{}>{\raggedright\arraybackslash}p{9em}>{\raggedright\arraybackslash}p{10em}rrlllr@{}}
    \toprule
    Benchmark & Pairs read & Pairs & Opposite & Random, & R1 & R2 & Reversals \\
     & & & signs & same size & & & \\
    \midrule
    SWE-bench Lite & kept part and parent & \val{276}{txt:results/subsets/summary.md@\n.. Case A[\s\S]*?Simultaneous 95. intervals, (\d+) pairs} & \val{4}{txt:results/subsets/summary.md@\n.. Case A[\s\S]*?uniform, central 95.\): tau_b [\d.]+ against a reference of [\d.]+ to [\d.]+ \(percentile [\d.]+\); opposite-sign pairs (\d+) against} & \val{1}{txt:results/subsets/summary.md@\n.. Case A[\s\S]*?uniform, central 95.\): tau_b [\d.]+ against a reference of [\d.]+ to [\d.]+ \(percentile [\d.]+\); opposite-sign pairs \d+ against a reference of (\d+) to}\textendash{}\val{9}{txt:results/subsets/summary.md@\n.. Case A[\s\S]*?uniform, central 95.\): tau_b [\d.]+ against a reference of [\d.]+ to [\d.]+ \(percentile [\d.]+\); opposite-sign pairs \d+ against a reference of \d+ to ([\d.]+) \(} & & & \val{0}{txt:results/subsets/summary.md@\n.. Case A[\s\S]*?Simultaneous 95. intervals, \d+ pairs: resolved on kept part \d+, on parent \d+; reversals (\d+)} \\
    SWE-bench Verified & kept part and parent & \val{276}{txt:results/subsets/summary.md@\n.. Case B[\s\S]*?Simultaneous 95. intervals, (\d+) pairs} & \val{2}{txt:results/subsets/summary.md@\n.. Case B[\s\S]*?uniform, central 95.\): tau_b [\d.]+ against a reference of [\d.]+ to [\d.]+ \(percentile [\d.]+\); opposite-sign pairs (\d+) against} & \val{0}{txt:results/subsets/summary.md@\n.. Case B[\s\S]*?uniform, central 95.\): tau_b [\d.]+ against a reference of [\d.]+ to [\d.]+ \(percentile [\d.]+\); opposite-sign pairs \d+ against a reference of (\d+) to}\textendash{}\val{6}{txt:results/subsets/summary.md@\n.. Case B[\s\S]*?uniform, central 95.\): tau_b [\d.]+ against a reference of [\d.]+ to [\d.]+ \(percentile [\d.]+\); opposite-sign pairs \d+ against a reference of \d+ to ([\d.]+) \(} & & & \val{0}{txt:results/subsets/summary.md@\n.. Case B[\s\S]*?Simultaneous 95. intervals, \d+ pairs: resolved on kept part \d+, on parent \d+; reversals (\d+)} \\
    BigCodeBench-Hard (Complete) & kept part and parent & \val{11,628}{txt:results/subsets/summary.md@\n.. Case C[\s\S]*?Simultaneous 95. intervals, (\d+) pairs} & \val{935}{txt:results/subsets/summary.md@\n.. Case C[\s\S]*?uniform, central 95.\): tau_b [\d.]+ against a reference of [\d.]+ to [\d.]+ \(percentile [\d.]+\); opposite-sign pairs (\d+) against} & \val{615}{txt:results/subsets/summary.md@\n.. Case C[\s\S]*?uniform, central 95.\): tau_b [\d.]+ against a reference of [\d.]+ to [\d.]+ \(percentile [\d.]+\); opposite-sign pairs \d+ against a reference of (\d+) to}\textendash{}\val{975}{txt:results/subsets/summary.md@\n.. Case C[\s\S]*?uniform, central 95.\): tau_b [\d.]+ against a reference of [\d.]+ to [\d.]+ \(percentile [\d.]+\); opposite-sign pairs \d+ against a reference of \d+ to ([\d.]+) \(} & & & \val{0}{txt:results/subsets/summary.md@\n.. Case C[\s\S]*?Simultaneous 95. intervals, \d+ pairs: resolved on kept part \d+, on parent \d+; reversals (\d+)} \\
    \addlinespace
    MMLU, kept by MMLU-Pro & filter and clean systems & \val{2,775}{txt:results/mmlu_reconstruction/summary_pairs.md@- 75 systems, (\d+) pairs, \d+ tied} & \val{261}{txt:results/mmlu_reconstruction/summary_pairs.md@Kendall tau_b between the scores on kept and on all questions [\d.]+; (\d+) of} & \val{16}{txt:results/mmlu_reconstruction/summary_pairs.md@\(261 against (\d+) to}\textendash{}\val{39}{txt:results/mmlu_reconstruction/summary_pairs.md@\(261 against \d+ to (\d+)\)} & & & \val{0}{txt:results/mmlu_reconstruction/summary_pairs.md@Simultaneous 95. intervals: resolved on kept \d+, on all \d+, on removed \d+; \*\*reversals (\d+)\*\*} \\
     & \quad both at or above the floor & \val{595}{txt:results/mmlu_reconstruction/summary_pairs_posthoc.md@\| A \| 30. \| \d+ \| (\d+) \|} & \val{7}{txt:results/mmlu_reconstruction/summary_pairs_posthoc.md@\| A \| 30. \| \d+ \| \d+ \| (\d+) \|} & \val{1}{txt:results/mmlu_reconstruction/summary_pairs_posthoc.md@\| A \| 30. \| \d+ \| \d+ \| \d+ \| (\d+) to}\textendash{}\val{8}{txt:results/mmlu_reconstruction/summary_pairs_posthoc.md@\| A \| 30. \| \d+ \| \d+ \| \d+ \| \d+ to (\d+) \|} & & & \val{0}{txt:results/mmlu_reconstruction/summary_pairs_posthoc.md@\| A \| 30. \| \d+ \| \d+ \| \d+ \| \d+ to \d+ \| [^|]*\| [^|]*\| \d+ / (\d+) \|} \\
     & with descendants (\val{154}{txt:results/mmlu_reconstruction/summary_pairs.md@Population K: the systems of known provenance[^\n]*\n- (\d+) systems} systems) & \val{11,781}{txt:results/mmlu_reconstruction/summary_pairs.md@- 154 systems, (\d+) pairs, \d+ tied} & \val{909}{txt:results/mmlu_reconstruction/summary_pairs.md@Population K: the systems of known provenance[\s\S]*?Kendall tau_b between the scores on kept and on all questions [\d.]+; (\d+) of} & & & & \val{36}{txt:results/mmlu_reconstruction/summary_pairs.md@\*\*reversals (\d+)\*\*; newly resolved 1194} \\
    \addlinespace
    MMLU, matched & clean--clean & \val{1,191}{txt:results/mmlu_reconstruction/summary_difficulty_reference.md@\| clean.clean, same half \| (\d+) \| \d+ \| [\d.]+ to [\d.]+ \([\d.]+\)} & \val{113}{txt:results/mmlu_reconstruction/summary_difficulty_reference.md@\| clean.clean, same half \| \d+ \| (\d+) \| [\d.]+ to [\d.]+ \([\d.]+\)} & \val{6}{txt:results/mmlu_reconstruction/summary_difficulty_reference.md@\| clean.clean, same half \| \d+ \| \d+ \| ([\d.]+) to [\d.]+ \(}\textendash{}\val{19}{txt:results/mmlu_reconstruction/summary_difficulty_reference.md@\| clean.clean, same half \| \d+ \| \d+ \| [\d.]+ to ([\d.]+) \(} & \val{28}{txt:results/mmlu_reconstruction/summary_difficulty_reference.md@\| clean.clean, same half \| \d+ \| \d+ \| [\d.]+ to [\d.]+ \([\d.]+\) \| ([\d.]+) to [\d.]+ \(}\textendash{}\val{48}{txt:results/mmlu_reconstruction/summary_difficulty_reference.md@\| clean.clean, same half \| \d+ \| \d+ \| [\d.]+ to [\d.]+ \([\d.]+\) \| [\d.]+ to ([\d.]+) \(} & \val{46}{txt:results/mmlu_reconstruction/summary_difficulty_reference.md@\| clean.clean, same half \| \d+ \| \d+ \| [\d.]+ to [\d.]+ \([\d.]+\) \| [\d.]+ to [\d.]+ \([\d.]+\) \| ([\d.]+) to [\d.]+ \(}\textendash{}\val{63}{txt:results/mmlu_reconstruction/summary_difficulty_reference.md@\| clean.clean, same half \| \d+ \| \d+ \| [\d.]+ to [\d.]+ \([\d.]+\) \| [\d.]+ to [\d.]+ \([\d.]+\) \| [\d.]+ to ([\d.]+) \(} & \\
    references & \quad both at or above the floor & \val{241}{txt:results/mmlu_reconstruction/summary_difficulty_reference.md@\| clean.clean, same half \| 0\.30 \| both above \| (\d+) \|} & \val{3}{txt:results/mmlu_reconstruction/summary_difficulty_reference.md@\| clean.clean, same half \| 0\.30 \| both above \| \d+ \| (\d+) \|} & \val{0}{txt:results/mmlu_reconstruction/summary_difficulty_reference.md@\| clean.clean, same half \| 0\.30 \| both above \| \d+ \| \d+ \| (\d+) to}\textendash{}\val{4}{txt:results/mmlu_reconstruction/summary_difficulty_reference.md@\| clean.clean, same half \| 0\.30 \| both above \| \d+ \| \d+ \| \d+ to (\d+) } & \val{1}{txt:results/mmlu_reconstruction/summary_difficulty_reference.md@\| clean.clean, same half \| 0\.30 \| both above \| \d+ \| \d+ \| [^|]*\| (\d+) to}\textendash{}\val{5}{txt:results/mmlu_reconstruction/summary_difficulty_reference.md@\| clean.clean, same half \| 0\.30 \| both above \| \d+ \| \d+ \| [^|]*\| \d+ to (\d+) } & \val{1}{txt:results/mmlu_reconstruction/summary_difficulty_reference.md@\| clean.clean, same half \| 0\.30 \| both above \| \d+ \| \d+ \| [^|]*\| [^|]*\| (\d+) to}\textendash{}\val{5}{txt:results/mmlu_reconstruction/summary_difficulty_reference.md@\| clean.clean, same half \| 0\.30 \| both above \| \d+ \| \d+ \| [^|]*\| [^|]*\| \d+ to (\d+) } & \\
     & filter--clean & \val{350}{txt:results/mmlu_reconstruction/summary_difficulty_reference.md@\| filter.clean \| (\d+) \| \d+ \| [\d.]+ to [\d.]+ \([\d.]+\)} & \val{28}{txt:results/mmlu_reconstruction/summary_difficulty_reference.md@\| filter.clean \| \d+ \| (\d+) \| [\d.]+ to [\d.]+ \([\d.]+\)} & \val{0}{txt:results/mmlu_reconstruction/summary_difficulty_reference.md@\| filter.clean \| \d+ \| \d+ \| ([\d.]+) to [\d.]+ \(}\textendash{}\val{4}{txt:results/mmlu_reconstruction/summary_difficulty_reference.md@\| filter.clean \| \d+ \| \d+ \| [\d.]+ to ([\d.]+) \(} & \val{0}{txt:results/mmlu_reconstruction/summary_difficulty_reference.md@\| filter.clean \| \d+ \| \d+ \| [\d.]+ to [\d.]+ \([\d.]+\) \| ([\d.]+) to [\d.]+ \(}\textendash{}\val{5}{txt:results/mmlu_reconstruction/summary_difficulty_reference.md@\| filter.clean \| \d+ \| \d+ \| [\d.]+ to [\d.]+ \([\d.]+\) \| [\d.]+ to ([\d.]+) \(} & \val{0}{txt:results/mmlu_reconstruction/summary_difficulty_reference.md@\| filter.clean \| \d+ \| \d+ \| [\d.]+ to [\d.]+ \([\d.]+\) \| [\d.]+ to [\d.]+ \([\d.]+\) \| ([\d.]+) to [\d.]+ \(}\textendash{}\val{4}{txt:results/mmlu_reconstruction/summary_difficulty_reference.md@\| filter.clean \| \d+ \| \d+ \| [\d.]+ to [\d.]+ \([\d.]+\) \| [\d.]+ to [\d.]+ \([\d.]+\) \| [\d.]+ to ([\d.]+) \(} & \\
     & descendant--clean & \val{5,530}{txt:results/mmlu_reconstruction/summary_difficulty_reference.md@\| lineage.clean \| (\d+) \| \d+ \| [\d.]+ to [\d.]+ \([\d.]+\)} & \val{375}{txt:results/mmlu_reconstruction/summary_difficulty_reference.md@\| lineage.clean \| \d+ \| (\d+) \| [\d.]+ to [\d.]+ \([\d.]+\)} & \val{8}{txt:results/mmlu_reconstruction/summary_difficulty_reference.md@\| lineage.clean \| \d+ \| \d+ \| ([\d.]+) to [\d.]+ \(}\textendash{}\val{25}{txt:results/mmlu_reconstruction/summary_difficulty_reference.md@\| lineage.clean \| \d+ \| \d+ \| [\d.]+ to ([\d.]+) \(} & \val{24}{txt:results/mmlu_reconstruction/summary_difficulty_reference.md@\| lineage.clean \| \d+ \| \d+ \| [\d.]+ to [\d.]+ \([\d.]+\) \| ([\d.]+) to [\d.]+ \(}\textendash{}\val{42}{txt:results/mmlu_reconstruction/summary_difficulty_reference.md@\| lineage.clean \| \d+ \| \d+ \| [\d.]+ to [\d.]+ \([\d.]+\) \| [\d.]+ to ([\d.]+) \(} & \val{58}{txt:results/mmlu_reconstruction/summary_difficulty_reference.md@\| lineage.clean \| \d+ \| \d+ \| [\d.]+ to [\d.]+ \([\d.]+\) \| [\d.]+ to [\d.]+ \([\d.]+\) \| ([\d.]+) to [\d.]+ \(}\textendash{}\val{73}{txt:results/mmlu_reconstruction/summary_difficulty_reference.md@\| lineage.clean \| \d+ \| \d+ \| [\d.]+ to [\d.]+ \([\d.]+\) \| [\d.]+ to [\d.]+ \([\d.]+\) \| [\d.]+ to ([\d.]+) \(} & \\
     & \quad Llama-2-7b descendants & \val{770}{txt:results/mmlu_reconstruction/summary_difficulty_reference.md@\| Llama-2-7b \| (\d+) \| \d+ \| [\d.]+ to [\d.]+ \([\d.]+\)} & \val{315}{txt:results/mmlu_reconstruction/summary_difficulty_reference.md@\| Llama-2-7b \| \d+ \| (\d+) \| [\d.]+ to [\d.]+ \([\d.]+\)} & \val{1}{txt:results/mmlu_reconstruction/summary_difficulty_reference.md@\| Llama-2-7b \| \d+ \| \d+ \| ([\d.]+) to [\d.]+ \(}\textendash{}\val{9}{txt:results/mmlu_reconstruction/summary_difficulty_reference.md@\| Llama-2-7b \| \d+ \| \d+ \| [\d.]+ to ([\d.]+) \(} & \val{6}{txt:results/mmlu_reconstruction/summary_difficulty_reference.md@\| Llama-2-7b \| \d+ \| \d+ \| [\d.]+ to [\d.]+ \([\d.]+\) \| ([\d.]+) to [\d.]+ \(}\textendash{}\val{18}{txt:results/mmlu_reconstruction/summary_difficulty_reference.md@\| Llama-2-7b \| \d+ \| \d+ \| [\d.]+ to [\d.]+ \([\d.]+\) \| [\d.]+ to ([\d.]+) \(} & \val{34}{txt:results/mmlu_reconstruction/summary_difficulty_reference.md@\| Llama-2-7b \| \d+ \| \d+ \| [\d.]+ to [\d.]+ \([\d.]+\) \| [\d.]+ to [\d.]+ \([\d.]+\) \| ([\d.]+) to [\d.]+ \(}\textendash{}\val{44}{txt:results/mmlu_reconstruction/summary_difficulty_reference.md@\| Llama-2-7b \| \d+ \| \d+ \| [\d.]+ to [\d.]+ \([\d.]+\) \| [\d.]+ to [\d.]+ \([\d.]+\) \| [\d.]+ to ([\d.]+) \(} & \\
     & descendant--descendant or filter & \val{3,476}{txt:results/mmlu_reconstruction/summary_difficulty_reference.md@\| lineage.lineage and lineage.filter \| (\d+) \| \d+ \| [\d.]+ to [\d.]+ \([\d.]+\)} & \val{273}{txt:results/mmlu_reconstruction/summary_difficulty_reference.md@\| lineage.lineage and lineage.filter \| \d+ \| (\d+) \| [\d.]+ to [\d.]+ \([\d.]+\)} & \val{47}{txt:results/mmlu_reconstruction/summary_difficulty_reference.md@\| lineage.lineage and lineage.filter \| \d+ \| \d+ \| ([\d.]+) to [\d.]+ \(}\textendash{}\val{94}{txt:results/mmlu_reconstruction/summary_difficulty_reference.md@\| lineage.lineage and lineage.filter \| \d+ \| \d+ \| [\d.]+ to ([\d.]+) \(} & \val{90}{txt:results/mmlu_reconstruction/summary_difficulty_reference.md@\| lineage.lineage and lineage.filter \| \d+ \| \d+ \| [\d.]+ to [\d.]+ \([\d.]+\) \| ([\d.]+) to [\d.]+ \(}\textendash{}\val{119.5}{txt:results/mmlu_reconstruction/summary_difficulty_reference.md@\| lineage.lineage and lineage.filter \| \d+ \| \d+ \| [\d.]+ to [\d.]+ \([\d.]+\) \| [\d.]+ to ([\d.]+) \(} & \val{132}{txt:results/mmlu_reconstruction/summary_difficulty_reference.md@\| lineage.lineage and lineage.filter \| \d+ \| \d+ \| [\d.]+ to [\d.]+ \([\d.]+\) \| [\d.]+ to [\d.]+ \([\d.]+\) \| ([\d.]+) to [\d.]+ \(}\textendash{}\val{164.5}{txt:results/mmlu_reconstruction/summary_difficulty_reference.md@\| lineage.lineage and lineage.filter \| \d+ \| \d+ \| [\d.]+ to [\d.]+ \([\d.]+\) \| [\d.]+ to [\d.]+ \([\d.]+\) \| [\d.]+ to ([\d.]+) \(} & \\
    \bottomrule
  \end{tabular}
\end{table}

\subsection*{ScopeCard for diagnosis and reproduction}

ScopeCard turns these decision rules into a check that a benchmark's user can run. From a release, a declared
universe and the evaluation protocol, it answers four questions: whether a ranking claim on the benchmark can be
read as a claim about the target at all (the regime); how much of each test query's candidates the benchmark
excludes ($W_q$); how large an unmeasured change between the regions would have to be to reverse a reported
difference ($c^*$, only for a metric that decomposes); and whether the stated construction rule reproduces the
release. It maps the release's ids onto the universe, stops when most do not match, and writes a versioned scope
card that states the regime and only what the regime supports. The construction rule is diagnosed, never assumed:
four families of depth rules are searched over thresholds, and the best of each is reported with its Jaccard
similarity $J$ to the release, as reproduced at $J = 1$ or approximately reproduced at
$J \ge \val{0.95}{src:design constant, tool/src/scopecard/rules.py}$. One package, needing only NumPy, runs
from the command line and, through Pyodide, in a browser page that keeps the files on the user's machine.

Three checks carry the tool's validation (Extended Data Table~\ref{ed:validation}). First, the answers it is checked
against come from a separate plain implementation, which agrees with the tool's on all
\val{2,400}{txt:results/validation/v2_summary.md@select the same units for (\d+) of} family--threshold pairs
tested; on synthetic releases built by each rule family, the search returns the generating units in every
noise-free run, the verdict is reproduced or approximately reproduced in at least
\val{98.5}{calc:{txt:results/validation/v2_summary.md@\| 0\.02 \| \d+ \| \S+ \| \S+ \| ([\d.]+)}+{txt:results/validation/v2_summary.md@\| 0\.02 \| \d+ \| \S+ \| \S+ \| \S+ \| ([\d.]+)}}\%
of runs when up to a share $\epsilon = \val{0.02}{src:design constant, scripts/validate_synthetic.py NOISE}$ of the
units is moved at random, in none from $\epsilon = \val{0.05}{src:design constant, scripts/validate_synthetic.py NOISE}$,
and never for random or out-of-family control releases. Second, the tool refuses or warns where the files do not
support a claim: all \val{27}{txt:results/validation/v3_refusals.md@\d+ of (\d+) scenarios behave} scenarios of
mismatched, malformed, empty or contradictory inputs behave as specified. Third, one command reconstructs the
targets, numerical analyses and scope cards from public inputs and the supplied per-query scores, without
retraining, and checks them against the stored results in a new environment. The user tests have so far been
internal.

For filters that use model outputs, the analyses of the two previous sections are released as scripts in the
reproduction package. Their inputs indicate what a release of such a benchmark should report: the panel's members and
their lines, the rule and its handling of ties, the size of the panel, and whether any scored outputs are outputs that
voted. Extended Data Table~\ref{ed:claims} lists, for each conclusion of this paper, the inputs it rests on, the
checks it passed and what carries it: a field of the current scope card, a script, or one of these recommended
fields.

\section*{Discussion}

The two kinds of construction rule call for different checks. Under a fixed rule, the audit determines whether the
released files identify the excluded share and, when they do, measures it. Whether a comparison reverses on the target
needs labels on the excluded part; on WN18RR, where they exist, the answer depended on the metric and on the target's
degree composition.
Under a filter that uses model outputs, the kept set depends on the panel: a member's own vote cannot raise its score,
and on BigCodeBench the shortfall also reached outputs that did not vote and, in a panel concentrated in one line,
that line's non-voting models. On the MMLU questions that MMLU-Pro retained, the order changed more than matched
references give, mainly near the guessing floor and in the Llama-2 line.

Measurements of excluded parts in two biological classification settings point the same way. In
post-translational-modification site prediction, a model improvement that holds on the benchmark's constructed
negatives reverses on reconstructed negatives in seven of eight tasks\citep{li2026constructed}; in drug--target
interaction, AUROC is lower for every model on the negatives that an eligibility rule excludes\citep{li2026lowbias}.
WN18RR adds a ranking benchmark, where the metric matters as well as the target.

These results suggest four practices for benchmark releases and ranking claims.
\begin{enumerate}
  \item \emph{State the population and the comparison.} A release should declare its universe and evaluated unit,
    and a ranking claim should name its target population and its metric. Datasheets ask whether a dataset is a
    sample of a larger set and whether it is representative\citep{gebru2021datasheets}, and data cards summarise
    provenance and intended use\citep{pushkarna2022data}; a scope card answers the first question with a number, at
    the evaluated unit.
  \item \emph{Make the selection reconstructible.} A release should state a construction rule that reproduces it;
    where a filter uses model outputs, it should also name the panel and the lines of its members, state the rule and
    how it treats ties, and keep the panel's item-level outcomes.
  \item \emph{Report the diagnostics that the evidence supports.} For a fixed rule, these are the regime and $W$, per
    query where candidates are enumerated, and $c^*$ for a metric that decomposes, with $\Delta_O$ measured on the
    target's queries under its filter. On a benchmark filtered by model outputs, a comparison should say whether a
    compared model, or a model of its line, sat on the panel, report those models apart from the others, and say
    whether the scored outputs are the outputs that voted: when they are, the penalty for taking part has a known
    sign. For other outputs, the direction must be measured.
  \item \emph{Make the comparison auditable.} Keep the selection decisions, the input versions, the item-level
    outcomes and the analysis code needed to reproduce the comparison, and report changes in score, sign changes and
    established reversals separately.
\end{enumerate}

Each remedy at hand addresses part of the problem. In the panels studied, a larger panel reduced a member's penalty
for its own vote, but not the part added by the other builders of its line. Choosing the items with one half of the
model owners and scoring the other half\citep{zheng2026neartied} removes a model's own participation by
construction; whether related models released by other owners carry part of the effect is open, because the
composition study grouped models by line, not by owner. Spreading a panel across lines reduced the concentrated
line's disadvantage; whether it gives a better ranking was not tested. Where the excluded part of a
fixed rule is not observable, labelling a sample of it can settle a comparison, and $c^*$ states how large a change
the labels must rule out. The machinery exists: active comparison\citep{sawade2012active} and active
testing\citep{kossen2021active}, and in retrieval, choosing documents to judge until the sign of a difference is
settled\citep{carterette2006minimal} or topics for a given power\citep{sakai2016topic}. It needs both models to score
the sampled units. Retrieval systems and feature-based models can; transductive embeddings and identifier-based
recommenders cannot score units they never saw, so their excluded part is measured by training on the target, as on
WN18RR.

The study has limits. One ranking benchmark is measured, and the ten audited benchmarks were chosen by reuse, not
sampled. Candidates follow the filtered convention. $W$ counts different units in different benchmarks, so its
values are not comparable between them, and three universes are unavailable. Per-pair intervals on WN18RR are not
corrected for multiple comparisons, and NBFNet enters only as a deployment test. $c^*$ is exact only for a $\Delta_O$
measured on the same queries under the same filter, and carries no guarantee otherwise. The studies of filters that
use model outputs are exploratory and rest on archived item-level runs, one per model and prompt format; their panels
were drawn at random within the published rules, whereas builders choose panels; model lines are read from names; and
the panel experiments used only BigCodeBench and MMLU. The two prompt formats share models and tasks, so the cross-format result is
not an independent rerun. Five of MMLU-Pro's eight filter models have archived MMLU runs, and the questions that
MMLU-Pro retained are identified by their text; they are not a replay of the filter and cannot separate the panel's
choice from MMLU-Pro's later review. The difficulty-matched references were
tuned on the scored systems' levels and leave the matched subsets slightly easier than the kept set. The user tests of
the tool have so far been internal.

\clearpage
\section*{Methods}

\subsection*{Notation and identification}

Extended Data Table~\ref{ed:notation} lists the symbols of the Results and Methods. Throughout, $Q$ is a finite set
of test queries, and means and covariances over $Q$ are unweighted population moments. For a candidate set $S$ and a
scoring function $s$, write
\[
  b_q(S) = \sum_{v\in S}\Bigl(\mathbf{1}[s(v)<s(p_q)] + \tfrac12\,\mathbf{1}[s(v)=s(p_q)]\Bigr),
  \qquad \mathrm{AUC}_q(S) = b_q(S)/|S| ,
\]
so that ties at the positive's score count one half. For two models $A$ and $B$, $\Delta_{O,q}$, $\Delta_{U,q}$ and
$\Delta_{C,q}$ are the differences $A-B$ in $\mathrm{AUC}_q$ on $O_q$, $U_q$ and $C_q$, and $\Delta_O$, $\Delta_U$
and $\Delta_{\mathrm{target}}$ are their means over $Q$. If candidates carry weights, as in a target re-weighted to
the benchmark's degree profile, every count becomes a sum of weights and the arguments below are unchanged.

\begin{proposition}[Decomposition]\label{prop:nmi-decomp}
(a) If exclusion removes a share $0<W<1$ of the queries, then for any metric averaged over queries,
$\Delta_{\mathrm{target}} = (1-W)\Delta_O + W\Delta_U$.
(b) If exclusion removes candidates within queries, and every query has both kept and excluded candidates, then for
per-query AUC, $\Delta_{C,q} = (1-W_q)\Delta_{O,q} + W_q\Delta_{U,q}$ for every query, and
$\Delta_{\mathrm{target}} = \Delta_O - \bar W\,\bar c - \operatorname{Cov}_q(W_q, c_q)$, where
$c_q = \Delta_{O,q} - \Delta_{U,q}$ and $\bar c = \Delta_O - \Delta_U$.
\end{proposition}

\begin{proof}
(a) With $Q_O$ the kept and $Q_U$ the excluded queries, $|Q_U| = W|Q|$, the mean of any per-query metric over
$Q$ is $(1-W)$ times its mean over $Q_O$ plus $W$ times its mean over $Q_U$; subtracting the two models gives the
claim. (b) $b_q$ is additive over disjoint sets, so $b_q(C_q) = b_q(O_q) + b_q(U_q)$; dividing by
$|C_q| = |O_q| + |U_q|$ gives $\mathrm{AUC}_q(C_q) = (1-W_q)\,\mathrm{AUC}_q(O_q) + W_q\,\mathrm{AUC}_q(U_q)$.
Subtracting the two models gives $\Delta_{C,q} = \Delta_{O,q} - W_q c_q$, and averaging over $Q$ gives
$\Delta_{\mathrm{target}} = \Delta_O - \overline{W_q c_q} = \Delta_O - \bar W\bar c - \operatorname{Cov}_q(W_q,c_q)$.
\end{proof}

Part (b) is the per-query form of the identity of \citet{li2026constructed}, which pools AUROC over
positive--negative pairs with a fixed set of positives. The covariance records how the excluded share varies with
the regional change; for ConvE $-$ DistMult on WN18RR,
$\operatorname{Cov}_q(W_q,c_q) = \val{7.4}{calc:10000000*{v:E2/primary/conve-distmult/mean/approx_err|neg}}\times10^{-7}$.

\begin{corollary}[Break-even]\label{cor:nmi-breakeven}
Let $\Delta_O\neq 0$, $\bar W>0$ and $c = \bar c + \operatorname{Cov}_q(W_q,c_q)/\bar W$, the $W_q$-weighted mean of the
$c_q$ (in part (a), or when $W_q$ is constant, $c = \bar c$ and $\bar W = W$). The target difference has the
opposite sign to $\Delta_O$ exactly when $c\cdot\operatorname{sign}(\Delta_O) > c^* = |\Delta_O|/\bar W$.
\end{corollary}

\begin{proof}
Proposition~\ref{prop:nmi-decomp} reads $\Delta_{\mathrm{target}} = \Delta_O - \bar W c$. Multiplying by
$\operatorname{sign}(\Delta_O)$ gives $|\Delta_O| - \bar W\,c\,\operatorname{sign}(\Delta_O)$, which is negative
exactly when $c\,\operatorname{sign}(\Delta_O) > c^*$, and zero at equality.
\end{proof}

The threshold applies to the change between the regions, not to the difference on the excluded part: with a
common $W$, the ordering reverses exactly when $\Delta_U\operatorname{sign}(\Delta_O) < -(1-W)|\Delta_O|/W$, a
smaller threshold. \citet{li2026constructed} state $c^*$ for the magnitude $|c|$, which gives a sufficient
condition for the sign to persist; with $c$ signed, the condition is necessary and sufficient. The identity holds
for a $\Delta_O$ computed on the target's test queries, each ranked against its kept candidates under the
target's filter. For a difference taken from a leaderboard, computed under the benchmark's own filter or on a
different set of test triples, the identity does not hold and $c^*$ carries no guarantee: without a bound on the
mismatch, the threshold has no controlled error, and $\Delta_O$ has to be recomputed on the target's queries
under its filter.

\paragraph{No ordering without assumptions.} If nothing is known about the excluded candidates, each
$\Delta_{U,q}$ can attain either endpoint of $[-1,1]$: placing every excluded candidate of $q$ below the positive for
one model and above it for the other leaves $\Delta_{O,q}$ unchanged. Then $\Delta_{\mathrm{target}}$ lies in
$[x-\bar W, x+\bar W]$ and attains both ends, with $x = \overline{(1-W_q)\Delta_{O,q}}$, and since
$|x|\le 1-\bar W$, the two ends have opposite signs whenever $\bar W > \tfrac12$. No strict ordering of
two models on the target is then identified, whatever $\Delta_O$ is. This is the per-query form of the
corresponding result of \citet{li2026constructed}.

\paragraph{Reciprocal rank and Hits@$k$.} Assume no ties, and let $a_q(S)$ be the number of candidates in $S$
scored above the positive. Then $a_q(C_q) = a_q(O_q) + a_q(U_q)$,
$\mathrm{RR}_q(S) = 1/(1+a_q(S))$ and $\mathrm{Hits@}k_q(S) = \mathbf{1}[a_q(S)<k]$, so neither metric is a
mixture of its regional values.

\begin{proposition}[Regional differences do not determine MRR or Hits@$k$]\label{prop:nmi-rr}
There are models $A$, $A'$ and $B$ and a set of queries, each with $W_q = \tfrac12$, such that $A$ and $A'$ have
the same positive mean reciprocal rank differences against $B$ on the kept and on the excluded candidates, yet $A$
is worse than $B$ on all candidates and $A'$ is better. The same holds for Hits@$k$.
\end{proposition}

\begin{proof}
Every query has three kept and three excluded candidates; the entries below are $(a_q(O_q), a_q(U_q))$.
For mean reciprocal rank, two queries:
\begin{center}
\small
\begin{tabular}{lcccccc}
\toprule
 & $q_1$ & $q_2$ & MRR on $O$ & MRR on $U$ & MRR on $C$ & AUC on $O$, $U$, $C$ \\
\midrule
Model $A$  & $(0,1)$ & $(1,0)$ & $\tfrac34$ & $\tfrac34$ & $\tfrac12$ & $\tfrac56,\ \tfrac56,\ \tfrac56$ \\
Model $A'$ & $(0,0)$ & $(1,1)$ & $\tfrac34$ & $\tfrac34$ & $\tfrac23$ & $\tfrac56,\ \tfrac56,\ \tfrac56$ \\
Model $B$  & $(0,0)$ & $(2,2)$ & $\tfrac23$ & $\tfrac23$ & $\tfrac35$ & $\tfrac23,\ \tfrac23,\ \tfrac23$ \\
\midrule
$A-B$      &         &         & $+\tfrac1{12}$ & $+\tfrac1{12}$ & $-\tfrac1{10}$ & $+\tfrac16,\ +\tfrac16,\ +\tfrac16$ \\
$A'-B$     &         &         & $+\tfrac1{12}$ & $+\tfrac1{12}$ & $+\tfrac1{15}$ & $+\tfrac16,\ +\tfrac16,\ +\tfrac16$ \\
\bottomrule
\end{tabular}
\end{center}
Model $A$ is better in MRR on the kept and on the excluded candidates and worse on all of them; $A'$ has the same
regional differences and is better. AUC behaves as Proposition~\ref{prop:nmi-decomp}(b) requires. For Hits@1,
four queries:
\begin{center}
\small
\begin{tabular}{lcccccc}
\toprule
 & $q_1$ & $q_2$ & $q_3$ & $q_4$ & Hits@1 on $O$, $U$, $C$ & difference against $B$ \\
\midrule
Model $A$  & $(0,1)$ & $(1,0)$ & $(0,1)$ & $(1,0)$ & $\tfrac12,\ \tfrac12,\ 0$ & $+\tfrac14,\ +\tfrac14,\ -\tfrac14$ \\
Model $A'$ & $(0,0)$ & $(0,0)$ & $(1,1)$ & $(1,1)$ & $\tfrac12,\ \tfrac12,\ \tfrac12$ & $+\tfrac14,\ +\tfrac14,\ +\tfrac14$ \\
Model $B$  & $(0,0)$ & $(1,1)$ & $(1,1)$ & $(1,1)$ & $\tfrac14,\ \tfrac14,\ \tfrac14$ & \\
\bottomrule
\end{tabular}
\end{center}
A query is a hit on $C$ only if it is a hit on both regions. For Hits@$k$, give each region $\max(3,k)$
candidates and replace every entry $1$ by $k$; all values are unchanged.
\end{proof}

A sufficient condition remains. Without ties, every query has
$\mathrm{RR}_q(O_q) + \mathrm{RR}_q(U_q) - 1 \le \mathrm{RR}_q(C_q) \le \mathrm{RR}_q(O_q)$: the upper bound
because $a_q(C_q)\ge a_q(O_q)$, and the lower bound because, with $x = \mathrm{RR}_q(O_q)$ and
$y = \mathrm{RR}_q(U_q)$, it reduces to $(x+y)(1-x)(1-y)\ge 0$ when $x+y>1$. Applying the lower bound to $A$ and
the upper bound to $B$ gives $\Delta_{\mathrm{target}} \ge \Delta_O + \Delta_U - 1$ in MRR, and the same holds for
Hits@1. For Hits@$k$ with $k\ge 2$ even this bound fails, since two regional hits can add up to a miss. When
exclusion removes whole queries, Proposition~\ref{prop:nmi-decomp}(a) covers MRR and Hits@$k$ as well.

\paragraph{Regimes.} Regime (i) needs three conditions: the protocol enumerates candidates or queries over the
release's units (filtered ranking over the entity set, ranking over all items, judged topics only, queries from
source papers only); the universe is available; and the release's ids map onto it. The first condition comes from
the protocol, not from the files, since a file format alone does not show that candidates are enumerated. In
regime (i$'$), a declared enumeration, for example that every pair among the release's units is evaluated, must be
supported by the files: a declaration they contradict yields no $c^*$, and one they cannot check is reported as
resting on the declaration. Regime (ii) reports coverage only, because absence is not exclusion.

\begin{remark}[Absence is not exclusion]\label{rem:nmi-absence}
For a release $R\subset V$, a rule that keeps exactly $R$ and a sampler that includes every unit independently
with a probability $\pi\in(0,1)$ both give $R$ positive probability. The excluded share is $|V\setminus R|/|V|$
under the first and zero under the second, so one release cannot tell them apart; its coverage $|R|/|V|$ is the
same under both.
\end{remark}

\subsection*{ScopeCard}

\paragraph{Inputs and checks.} A release is read in the format of its task type: triple files for link
prediction, the LightGCN split format for recommendation, judgment files for retrieval, and pair files for pair
classification. The audit maps the release's ids onto the declared universe. It refuses a release when fewer than
half of its ids occur in the universe and warns when fewer than
\val{95}{src:design constant, tool/src/scopecard/mapping.py MAP_WARN}\% do; it refuses malformed lines,
naming the file and line, and empty files, naming the file; it counts the lines of a raw source that do not fit
its format and warns; it checks numeric inputs (a model difference in $[-1,1]$, a corpus at least as large as the
documents judged for any topic); and it checks a declared candidate rule against the files.

\paragraph{Per-query shares.} For filtered link prediction, a test query's candidates are the universe's units
minus the query's known answers, that is, the universe's triples of that relation, the release's triples and the
positive; $W_q$ is the share of them outside the release. For all-item recommendation, a test user's candidates
are the universe's items minus those the user is known to have, and users outside the release are queries
excluded as a whole, reported as a separate share. The card records the number of queries, the mean and spread of
$W_q$, and the queries without any candidate, without an excluded candidate or without a kept one. $c^*$ uses the
mean of $W_q$ and is withheld when a query keeps no candidate, since $\Delta_O$ is undefined there; when some test
triples fall outside the universe, the card warns that a model difference measured on all of them is on a
different query set, for which $c^*$ carries no guarantee.

\paragraph{Rule diagnosis.} Each family has one threshold $T$ on depth, searched over $T = 1,\dots,40$. A unit
threshold keeps the units of depth at least $T$; both ends keeps the ends of relations whose two ends both have
depth at least $T$; seeds with neighbours keeps the units of depth at least $T$ and everything they relate to; the
$k$-core keeps the largest sub-relation in which every unit has at least $T$ relations. Depth is a count in the
universe: triples per entity, references per paper, interactions per user or item, or judged-relevant passages per
topic. Release units that map onto no universe unit count against every rule. The best rule of each family is
reported with its Jaccard similarity to the release and a verdict, and a stated rule is checked the same way.

\paragraph{The scope card.} Besides the regime, $W$ and $c^*$, a card (schema
\val{0.3}{src:card schema version, tool/library/schema}) records the SHA-256 of every release file and a
fingerprint of the universe's content; the id mapping, with mapped, unmapped and merged ids; $W$ by stratum and, for pair universes, the
share of unobserved pairs and the release's pair coverage; the derived scope $d_{\min}$, the smallest depth at
which at most half of the units are excluded; hygiene checks (test triples whose reverse is a training triple,
test entities unseen in training, and for pair releases duplicate rows, self pairs and conflicting labels); and
the lines read and skipped in each input file. A card that records skipped lines but carries no warning fails
validation. Extended Data Table~\ref{ed:card} abridges the WN18RR card. The package needs only NumPy; the browser
page runs the same package in Pyodide and reads the files locally.

\subsection*{Validation}

V1 reruns the pilot's computations through the package: \val{10}{src:tool/tests/test_golden.py, number of golden tests}
golden tests read the pilot's input files and must reproduce every number of the pilot table for the
\val{7}{src:tool/library/benchmarks.json, benchmarks with a card} benchmarks with a card and for WN18. V2 builds
releases on \val{30}{txt:results/validation/v2_summary.md@On all (\d+) graphs} random heavy-tailed graphs by each
rule family, with answers from a separate plain-Python implementation, and moves a share $\epsilon$ of the units
in or out at random (Extended Data Table~\ref{ed:v2}); without noise, the search returns the generating units in
every run, in
\val{20.6}{calc:100-{txt:results/validation/v2_summary.md@\| 0 \| \d+ \| \S+ \| ([\d.]+)}}\% of them as a
different rule that selects the same units, which no release can tell apart. V3 is a suite of refusal and
warning scenarios, each a regression test (Extended Data Table~\ref{ed:v3}); every card written passes the schema
validation, and no scenario prints a traceback. V4 compares the browser page with the command line: their cards
are byte-identical, in JSON and Markdown, for WN18RR with and without a model difference, TREC DL 2019, Gowalla
and a pair-classification example. The check found two bugs, both fixed: a counting call that rejects 64-bit
integers on WebAssembly, and a Jaccard just below one that was rounded up to one beside the verdict
``approximately reproduced''. YAGO3-10, Amazon-Book and ogbl-citation2 exceed the browser's memory and are
audited from the command line only. V5 runs \texttt{reproduce.py} in a new environment: one tier runs the tests
that need no data; one rebuilds both declared WN18RR targets, checks them against recorded hashes, reruns the
prespecified analysis on the stored scorer outputs and compares every record; and one rebuilds every library
card, the example cards, the browser page and the card site from the public inputs and runs the full test suite.
GPU training and scoring are not rerun. V6 consists of two internal user tests of the browser page, whose
reported problems that affect what a card says are fixed; interface requests remain open. V7 records the cost
(Extended Data Table~\ref{ed:validation}). Further gaps found in the code are fixed and are now V3
scenarios: equal ids on the two sides of a bipartite universe read as self pairs, a pair count that is not a
share in a recommendation headline, two unchecked numeric inputs, and a break-even not tied to the queries and
filter it was computed on.

\subsection*{Ten benchmarks}

The anchors were fixed before any benchmark was inspected: in each domain, the benchmark set of one widely reused
study or benchmark suite. Each benchmark was coded from its papers, code and release for its domain, evaluated
unit, the source of its negatives or candidates, the rule stated in the paper, the rule found in code or release,
whether the text suffices to rebuild the release, and the declared universe. The construction rule was coded as
decidable and verified, decidable but unverified, partly decidable, or undecidable. Each benchmark with an
available universe was then audited by ScopeCard; the cards, with the SHA-256 of every release file and a
fingerprint of each universe, are part of the supplementary code, whose input list gives every input file with its
source and SHA-256. The prior-work status of the five findings was coded from the sources searched: WN18's entity
set and its merged ids are partly known; the entities FB15k-237 removed, YAGO3-10's both-ends rule and
ogbl-citation2's reference rule were not found in the sources searched; Amazon-Book's reduction of users is partly
known, from unanswered repository issues. For Amazon-Book, the stated 10-core reaches Jaccard
\val{0.502}{card:amazon-book@rule_diagnosis.stated.jaccard} with the release and the best $k$-core, with
$k = \val{14}{card:amazon-book@rule_diagnosis.best.T}$, reaches \val{0.726}{card:amazon-book@rule_diagnosis.best.jaccard}.
The TREC topic rule holds on all \val{43}{card:trec-dl-2019-passage@exclusion.included} judged topics of 2019 and
all \val{54}{card:trec-dl-2020-passage@exclusion.included} of 2020. Inside a judged topic, every passage left
unjudged counts as non-relevant: in 2019 that is at least
\val{0.99993}{card:trec-dl-2019-passage@exclusion.within_query.unjudged_share.min} of the corpus for every judged
topic. This is contamination of the candidates, which the pooling literature
studies\citep{zobel1998reliable,buckley2004retrieval}, not exclusion; exclusion in TREC acts on topics.

\subsection*{The WN18RR experiment}

\paragraph{Declared target.} Synsets are keyed by part of speech and offset. WN18RR's
\val{40,943}{card:wn18rr@mapping.release_ids} entity ids are bare offsets; mapped through the triples they stand
for, they give \val{41,105}{txt:results/wn18rr_target/summary.txt@\| O: (\d+)} synsets, because
\val{161}{card:wn18rr@mapping.merged_ids} ids each merge synsets of different parts of speech, one of them three (an adjective, a noun and a verb) and the
others two. Of WN18RR's
\val{93,003}{calc:{txt:results/wn18rr_target/summary.txt@WN18RR triples mapped: (\d+)}+{txt:results/wn18rr_target/summary.txt@unmatched: (\d+)}}
triples, \val{92,998}{txt:results/wn18rr_target/summary.txt@WN18RR triples mapped: (\d+)} map onto the target,
which has \val{218,367}{txt:results/wn18rr_target/summary.txt@target triples \(11 relations, synset level\) (\d+)}
triples. Two held-out samples of triples with at least one end in $U$, of
\val{3,034}{txt:MEASURED_C_DESIGN.md@`u_valid` ([\d,]+) and} and
\val{3,134}{txt:MEASURED_C_DESIGN.md@and `u_test` ([\d,]+) \(held-out} triples, are drawn so that both ends keep a
training triple. E2's training set is WN18RR's training triples together with every target triple outside the
held-out sets. Eighteen acceptance checks run after every build and all pass: entity ids and triples are unique;
every triple end is a target synset; the regions are exactly $O$ and $U$; the relations are WN18RR's 11 and no
others; for each WN18RR split, the mapped and the unmatched triples add up to the split; the mapped triples equal
WN18RR's minus the unmatched ones; $O$ equals the synsets of the mapped triples and covers exactly WN18RR's ids;
no held-out triple is in E2's training set; the training degrees match E2's training triples; each triple held
out for $U$ has an end in $U$; every $U$ candidate has an E2 training triple; $W$ recomputed from the entity table
matches the summary; the two samples have the sizes above; and a rebuild in a fresh folder is byte-identical. The
target of S3, keyed by bare offset as WN18RR keys it, has its own acceptance checks, which also pass.

\paragraph{Models and reproduction.} Each embedding model uses its unmodified WN18RR configuration from
\citet{ruffinelli2020you} in LibKGE\citep{broscheit2020libkge}, with five seeds, selected on the benchmark's
validation triples over $O$. In E1, the seed ranges of MRR contain the published values for all five, for
example \val{0.440}{min:E1/benchmark/conve/0,1,2,3,4/rr_O}\textendash{}\val{0.443}{max:E1/benchmark/conve/0,1,2,3,4/rr_O}
for ConvE against \val{0.442}{txt:MEASURED_C_DESIGN.md@`wnrr-conve.yaml` \| ([0-9.]+)} and
\val{0.450}{min:E1/benchmark/distmult/0,1,2,3,4/rr_O}\textendash{}\val{0.455}{max:E1/benchmark/distmult/0,1,2,3,4/rr_O}
for DistMult against \val{0.452}{txt:MEASURED_C_DESIGN.md@`wnrr-distmult.yaml` \| ([0-9.]+)}. NBFNet uses its
official code and configuration; a run on the target graph was estimated at
\val{50}{txt:MEASURED_C_DESIGN.md@roughly (\d+)–\d+ GPU-hours}\textendash{}\val{56}{txt:MEASURED_C_DESIGN.md@roughly \d+–(\d+) GPU-hours}
GPU-hours, so it was trained in E1 only and used for the deployment test. The popularity baseline scores a
candidate by how often it fills the query's relation slot in training. One counting module, which mirrors
LibKGE's treatment of ties, turns every model's scores into per-query counts; for the embedding models it
reproduces LibKGE's own scorer byte for byte.

\paragraph{Estimands and decision rules.} The primary estimand is per-query AUC on the benchmark test queries,
with MRR and Hits@$k$ reported alongside. S1 keeps only candidates whose part of speech fits the relation slot; S2
re-weights $U$ candidates within each query to the training-degree distribution of $O$, over the degree bins in
which the query has $U$ candidates and with $W_q$ unchanged, so that the synsets of $O$ without a training triple,
which have no counterpart in $U$, drop out of the weights; S3 keys synsets by bare
offset, as WN18RR does, and trains separately; S4 keeps in $O$ only synsets with a training triple. A difference
is resolved when its interval excludes zero, and a reversal is a pair whose $\Delta_O$ and
$\Delta_{\mathrm{target}}$ are both resolved with opposite signs. The prespecified rule names a reversal that
vanishes under S2 a rarity effect. A secondary, query-side estimand scores the held-out triples with an end in
$U$; it changes the positive reference and never enters $W$ or $c$ (Extended Data Table~\ref{ed:query-side}).

\paragraph{Deviations from the design.} The design (in the supplementary code) was prespecified before any model
was trained on the target; it was fixed in an internal, dated document, not in a public registry. It records four
amendments, each with when it was decided, and labels the post hoc analyses as such. NBFNet's scope was reduced to
E1 and the deployment test, which leaves 15 pairs instead of 21; no NBFNet result on the target existed then. The
planned mean number of excluded candidates in the top 10 cannot be computed from the stored counts, so the share
of queries outranked by at least one excluded candidate and the share that drop out of the top 10 are reported
instead; when this was decided, no comparison between $O$ and $U$ had been computed. Post hoc analyses were added
after the first E2 results showed queries between excluded synsets with an MRR of about
\val{0.85}{txt:MEASURED_C_DESIGN.md@U→U queries had MRR about (0\.\d+)}: a test triple leaks when its reverse is
an E2 training triple, which holds for \val{2,172}{v:E2/posthoc_bench_test:leak/complex/0/auc_C.n} of the
\val{6,266}{v:E2/posthoc_bench_test:all/complex/0/auc_C.n} benchmark test queries, and the primary estimand was
also computed on the \val{4,094}{v:E2/posthoc_noleak/complex/mean/auc_O.n} non-leaking ones. The deployment test's
analysis was specified before any trained NBFNet model was scored. Three clarifications added after the results
change no number: E1 and E2 differ on $O$ in training graph, synset identity and filter together, so their
difference is reported as a training/protocol effect (Extended Data Table~\ref{ed:protocol}); a result that
vanishes under S2 shows sensitivity to degree composition, not a mechanism; and no correction for multiple
comparisons was prespecified.

\paragraph{Deployment test.} Each NBFNet checkpoint trained in E1 reads E2's training graph at inference and
scores the benchmark test queries against $O$ and $U$ candidates under the target filter. Its graph-side effect,
deployment minus E1 on $O$ for the same checkpoint, changes the graph read at inference, the filter and the keying
of $O$ together (Extended Data Table~\ref{ed:deploy}). Per-model results under the primary estimand are in
Extended Data Table~\ref{ed:models}.

\subsection*{Statistics of the WN18RR experiment}

Every interval in the WN18RR experiment is a \val{95}{src:design constant, MEASURED_C_DESIGN.md Sect. 8}\% percentile interval from
\val{2,000}{txt:scripts/analyze_wn18rr.py@'--reps', type=int, default=(\d+)} bootstrap replicates over test
triples, in which both queries of a triple move together, or over head synsets in the clustered sensitivity
analysis. Intervals are per pair and per estimand, unadjusted for the 15 pairs and six estimands, and cover the
resampling of test triples. For a seed mean, the per-query metrics are averaged over the five trained seeds
before triples are resampled, so the interval conditions on these trained models and does not quantify the
uncertainty from training seeds; each seed is also reported on its own. The studies of filters that use model outputs
resample tasks or questions instead; their intervals are described with each study and listed in
Table~\ref{tab:intervals}.

\subsection*{Compute}

Audits run on a laptop CPU. Training and scoring used LibKGE at commit \texttt{9bca31f} and NBFNet-PyG at commit
\texttt{0c2d262}, each with a small patch, under PyTorch~2.4.0 with
CUDA~\val{12.4}{src:software version, not a result}. E1 seed~0 of the five embedding models ran on a laptop GPU;
all other training, and the scoring of those runs, used one A100 (80\,GB) or L40S GPU of a shared cluster. The
experiment took \val{187.48}{txt:results/compute/gpu_hours.md@\| total \| both \| \d+ \| ([\d.]+) \|} GPU-hours
in all (Extended Data Table~\ref{ed:compute}).

\subsection*{Filters that use model outputs}

\paragraph{The own-vote identity.} A panel $P$ of models votes on a pool of items; $y_{ji}\in\{0,1\}$ is the outcome
of model $j$ on item $i$ in the outputs that vote, and a threshold rule keeps $i$ when the number of members that pass
it is below a threshold (BigCodeBench-Hard) or at most a threshold (MMLU-Pro). Let $m\in P$, $u\notin P$,
$P' = (P\setminus\{m\})\cup\{u\}$, and let $H$ and $H'$ be the kept sets of $P$ and $P'$. The two counts on an item
differ by $y_{mi}-y_{ui}$, so the kept status changes only on items where the other members sit at the threshold and
$y_{mi}\neq y_{ui}$: the items of $E = H\setminus H'$, kept only when $m$ votes, have $y_{mi}=0$, and those of
$L = H'\setminus H$, kept only when $u$ votes, have $y_{mi}=1$. For a scoring vector $z$ over the items, let
$\bar z_X$ be its mean over a non-empty set $X$ of items, and write $e = |E|$ and $l = |L|$. For non-empty $H$
and $H'$, summing both scores over $H\cap H'$ gives
\begin{equation}
  \bar z_{H'} - \bar z_H = \frac{e\,(\bar z_{H'} - \bar z_E) + l\,(\bar z_L - \bar z_{H'})}{|H|}.
  \label{eq:filter-identity}
\end{equation}
where a term whose set is empty ($e = 0$ or $l = 0$) is zero. For the outputs that voted, $z = y_m$, so $\bar z_E = 0$ and $\bar z_L = 1$, and
\begin{equation}
  \bar y_{m,H'} - \bar y_{m,H} = \frac{e\,\bar y_{m,H'} + l\,(1-\bar y_{m,H'})}{|H|} \;\ge\; 0
  \label{eq:own-vote}
\end{equation}
whatever the panel, the threshold and the pool: a member's own vote cannot raise its score on the kept set. For
another scoring vector, such as $m$'s outcomes in another prompt format or a related model's outcomes,
equation~\eqref{eq:filter-identity} holds with that vector's means on $E$ and $L$, and its sign is not fixed: it is
positive when the vector tends to fail where $m$ fails (low $\bar z_E$) and to pass where $m$ passes (high $\bar z_L$).
The studies below average such differences over random panels and members, each against a reference in which another
member is replaced by the same outside model instead. The estimate is the difference between the two; it is not
guaranteed to be free of changes in the kept set's difficulty or content that the two replacements cause.

\begingroup\emergencystretch=2em
\paragraph{Data.} All inputs are public, read only and checked against stored hashes; no model was run. BigCodeBench\citep{zhuo2025bigcodebench}
v0.1.4 has \val{1,140}{txt:PANEL_COMPOSITION_DESIGN.md@BigCodeBench v0\.1\.4, ([\d,]+) tasks} tasks, with per-task
pass or fail for two prompt formats, Complete (a detailed docstring) and Instruct (a shorter natural-language
instruction), in the public tables of \texttt{bigcode/bigcodebench-perf} at one revision; 124 models are named
identically in both. For MMLU\citep{hendrycks2021measuring}, per-question results on the original four-option
questions (five-shot, one harness) come from the archive \texttt{open-llm-leaderboard-old}\citep{beeching2023open} on
Hugging Face: 268 systems, of which 5 are MMLU-Pro filter
models, 70 form a clean pool of systems not derived from the filter models (8 of them siblings, of a filter model's
family but not derived from it), 79 a lineage pool of systems derived from the filter models and 114 a random pool.
The repositories of the two Llama-2 chat filter models hold no MMLU records, and Yi-6B-Chat is not in the archive. An
original question is kept if MMLU-Pro holds a question of the same subject with the same text after lower-casing and
collapsing whitespace:
\val{6,580}{txt:results/mmlu_reconstruction/summary.md@same text and subject: (\d+) of} of
\val{14,042}{txt:results/mmlu_reconstruction/summary.md@same text and subject: \d+ of (\d+)}. The other questions were
removed by the filter or by MMLU-Pro's later review, which the release does not tell apart, and a question whose text
was changed counts as removed: the kept set is the set of questions retained in MMLU-Pro's release, not the output of
its filter alone. MMLU-Pro's own
per-question outputs come from its repository\citep{wang2024mmlupro}. Model lines are read from names by rules fixed before any outcome was
read; they are not verified against weights. Accuracies on MMLU are standardised to the kept questions' subject mix, over the subjects that have kept
questions.\par\endgroup

\paragraph{Panel size.} On each BigCodeBench table, and on MMLU over the 5 filter and 70 clean-pool systems under
MMLU-Pro's rule, 100 panels per size are drawn uniformly without replacement, at 4 to 96 builders on BigCodeBench and 4
to 64 on MMLU. Every member is evaluated when the panel has at most 16 builders, otherwise 16 drawn at random. For a
member $m$, a model $u$ outside the panel and outside $m$'s line is drawn; the penalty is $m$'s score on the kept set
of the panel with $m$ replaced by $u$, minus its score on the kept set of the panel with another member, not of $m$'s
line, replaced by $u$ instead. In a second contrast $m$ and the members of its line are replaced together. Intervals
come from a task or question bootstrap with 100 replicates, the panels and all random choices fixed.

\paragraph{Another prompt format.} The same random-panel design on the 124 models in both BigCodeBench tables, without
line exclusion: the panel votes with one format's outcomes, and the member is scored on the other format's outcomes,
which never vote. Ten models that vote in neither panel are scored on the same two kept sets, and the excess is the
member's change minus theirs. The primary contrast is Complete votes and Instruct scores at 8 builders; the other
sizes and the reverse direction are supplementary. A task bootstrap with
\val{2,000}{src:design constant, SEPARATED_SCORING_DESIGN.md} replicates is shared by both formats and all sizes;
simultaneous intervals use the bootstrap maximum of $|t|$ over the 20 contrasts of the curves. The design was fixed
before any cross-format score was computed.

\paragraph{Panel composition.} On BigCodeBench, every model line of at least two models is split once into an
evaluation pool, whose models never vote, and a builder pool; panel members are builder-pool models with a Complete
pass rate of at least
\val{25}{txt:PANEL_COMPOSITION_DESIGN.md@Complete pass rate of at least (\d+)\S*\sWithout}\%. For a focal line $\ell$
and $n$ builders, a concentrated panel holds $n/2$ members of $\ell$ and $n/2$ members of distinct other lines; its
spread partner keeps one member of $\ell$ and replaces each of the others by the eligible builder of a line not yet on
the panel nearest to it in Complete pass rate. Votes come from Complete outcomes and scores from Instruct outcomes. For
an evaluation model $w$, $d_w$ is its score on the spread panel's kept set minus its score on the concentrated panel's
kept set; its concentration cost is $d_w$ minus the mean of $d$ over its five controls, the evaluation models of
unchanged lines nearest to it in Instruct pass rate. $D(\ell,n)$ is the mean cost over the evaluation models of $\ell$
and over 100 panel pairs, and the primary estimand $D(8)$ is its mean over the seven focal lines at 8 builders, with
equal weight. The cost of spreading is the same quantity over the evaluation models of all lines that gained a seat,
pooled. Intervals come from a task bootstrap with
\val{1,000}{src:design constant, PANEL_COMPOSITION_DESIGN.md Section 7} replicates, percentile for $D(8)$ and
simultaneous (maximum of $|t|$) over the eight line contrasts. Before the run, constant, independent and
planted scoring vectors checked the code, and 40 scoring vectors drawn independently of the votes gave interval
coverage close to nominal; a check criterion found to be mis-specified was replaced before any Instruct outcome was
read (Deviation 1 of the design).

\paragraph{Penalty on the questions MMLU-Pro retained.} For a filter model $m$ with controls $G_m$, the clean-pool systems
whose accuracy on all questions lies within
\val{0.03}{txt:MMLU_RECONSTRUCTION_DESIGN.md@a\(all\) within ([\d.]+) of} of $m$'s (the five nearest if fewer), the
penalty is
\begin{equation}
  \delta_m = \bigl[\bar\alpha_{G_m}(\mathrm{kept}) - \alpha_m(\mathrm{kept})\bigr] - \bigl[\bar\alpha_{G_m}(\mathrm{all}) - \alpha_m(\mathrm{all})\bigr],
  \label{eq:mmlu-penalty}
\end{equation}
where $\alpha_m(\cdot)$ is $m$'s accuracy on a set of questions and $\bar\alpha_{G_m}(\cdot)$ the mean accuracy of its
controls, and $\bar\delta$ is the mean of $\delta_m$ over the five filter models. Intervals come from a bootstrap over
questions within subject
(\val{1,000}{src:design constant, MMLU_RECONSTRUCTION_DESIGN.md Section 4} replicates, control sets fixed). Every
clean-pool system inside the filter models' range of accuracy on all questions, widened by the same margin, receives a
placebo penalty with itself left out of its controls. The reference tail share is the share of subsets of this placebo
pool, of the size of the filter set, whose mean placebo penalty is at least $\bar\delta$; it locates $\bar\delta$ among
those subsets and is not a calibrated $p$-value, because the filter models are not exchangeable with the pool. On MMLU-Pro's own
questions (with up to ten options) the estimate has the same form, with MMLU-origin questions in place of the kept ones and
questions from other sources in place of all questions, but its design fixed other settings: six of the eight filter
models have outputs there; the controls of a filter model are the non-filter systems, siblings included, within
\val{0.10}{txt:MMLUPRO_FILTER_DESIGN.md@whose a\(O\) is within ([\d.]+) of} of its accuracy on the questions from
other sources (the three nearest if fewer); and the bootstrap resamples questions within category and source.

\paragraph{Order on the kept questions.} For each pair of systems, the differences in accuracy on the kept questions
and on all questions are compared. A pair changes sign when the two point differences have opposite signs; a reversal is
a pair whose two differences are resolved with opposite signs under simultaneous \val{95}{src:design constant, interval level}\% intervals (question bootstrap,
critical value from the maximum of $|t|$ over pairs). The reference is the central \val{95}{src:design constant, interval level}\% of the same count over
\val{2,000}{src:design constant, MMLU_RECONSTRUCTION_DESIGN.md Section 11} random subsets with the kept set's number of
questions per subject (R0). Pairs are read among the filter and clean-pool systems (75 systems), with the lineage pool
added (154), and among all systems (description only). A system is above the guessing floor at a given cut when its
accuracy on the kept questions is at least that cut; the cut of
\val{30}{src:post hoc cut, MMLU_RECONSTRUCTION_DESIGN.md Section 11}\%, with
\val{28}{src:post hoc cut, sensitivity}\% and \val{35}{src:post hoc cut, sensitivity}\%, was chosen after the run.

\paragraph{Difficulty-matched references.} The clean pool is split by model line into two halves. In each half, the
clean-pool systems of lines that hold no filter model and with at least
\val{40}{src:design constant, MMLU_RECONSTRUCTION_DESIGN.md Section 13}\% on all questions estimate each question's
difficulty, as the share of them
that answer it; the systems scored with these estimates are those of the other half, the filter models and the lineage
pool, so that no system is scored on strata its own outcomes helped to form. Within each subject, questions are cut into
eight strata of estimated difficulty. R1 draws, in every subject and stratum, as many questions as the kept set has
there; R2 also matches the answer key within strata. Each reference has 2,000 subsets per fold. Pairs of clean systems
from different halves are not evaluated. The estimators and the number of strata were chosen on the level gaps of the
scored systems, before any order statistic on a matched subset was computed, so the analysis is tuned on
evaluation-side levels and blind to the order statistics only. A synthetic check chose the kept set from the answers of
five synthetic filter models; it is therefore not a calibration under a null of difficulty alone, and only clear
excesses carry information. The matched subsets stayed up to about \val{2.0}{txt:MMLU_RECONSTRUCTION_DESIGN.md@stay up to about ([\d.]+) points \(R1\)} points (R1) and \val{1.8}{txt:MMLU_RECONSTRUCTION_DESIGN.md@stay up to about [\d.]+ points \(R1\) and ([\d.]+) points \(R2\)} points (R2)
easier than the kept set for the clean systems above \val{40}{src:bin edge of the floor description, summary_pairs_posthoc.md Section 1}\% on all questions; the effect of that gap
on the counts of sign changes has no known direction, and errors in the difficulty estimates, item discrimination,
content and MMLU-Pro's later review of the kept questions are not separated from the panel's choice.

\paragraph{Released subsets.} SWE-bench Lite\citep{jimenez2024swebench,jimenez2024swebenchlite} (a scope rule),
SWE-bench Verified\citep{chowdhury2024swebenchverified} (a human validity review) and
BigCodeBench-Hard\citep{zhuo2025bigcodebench} (a difficulty rule over model results) are compared with their parents on the same systems: per pair,
the differences on the kept part and on the parent, reversals under simultaneous \val{95}{src:design constant, interval level}\% intervals, and the number of
pairs with opposite signs against the central \val{95}{src:design constant, interval level}\% of random subsets of the parent of the kept part's size.

\paragraph{Readings and corrections.} Each design fixed its readings before its run. For the panel-size and
composition studies, an effect is \emph{material} if its interval excludes zero and the estimate is at least one point,
\emph{detectable but small} if the interval excludes zero and the estimate is below one point, and \emph{not detected}
otherwise, which is not evidence of absence. For the penalties on the questions MMLU-Pro retained and on MMLU-Pro's own
questions, the designs also required a reference tail share below
\val{0.05}{txt:MMLUPRO_FILTER_DESIGN.md@placebo p below ([\d.]+), and the bootstrap}; by that reading the first
penalty was confirmed and the second was not detected, although its interval excludes zero. Because a reference tail
share locates a value among reference subsets and is not a calibrated $p$-value, the Results report the estimate, its
interval and the tail share instead of these labels. The resampling of each study is listed in
Table~\ref{tab:intervals}. Deviations from each design are recorded in it with their dates. Of the corrections made on
2026-11-16 and 2026-11-17, one changed an estimate: after a run whose question identifiers did not match the questions
was excluded, the secondary regression of the MMLU-Pro analysis gave a mean shortfall of
\val{1.13}{txt:results/mmlupro_filter/summary.md@Mean shortfall of the filter models below the fitted curve: \+([\d.]+) points}
instead of \val{1.40}{txt:MMLUPRO_FILTER_DESIGN.md@shortfall of the filter models goes from ([\d.]+) to} points. The
others changed wording: a reference tail share is not a $p$-value; the cost of spreading is pooled over lines, and no
conservation and no equivalence of the kept sets were shown; and the share of an excess that a matched reference
reaches is descriptive and is not a share explained by difficulty or key.

\begin{table}[tbp]
  \caption{\textbf{Resampling in each study.} Every interval comes from a bootstrap over items; no study resamples
    systems or models, so the intervals reflect the sampling of items only. References are not intervals: the central
    \val{95}{src:design constant, interval level}\% of a count over random subsets of the kept part's size
    (\val{2,000}{src:design constant, MMLU_RECONSTRUCTION_DESIGN.md Section 11 and SUBSET_C_DESIGN.md} subsets) or over
    subsets also matched on estimated difficulty and answer key
    (\val{2,000}{src:design constant, MMLU_RECONSTRUCTION_DESIGN.md Section 13} per fold), and the tail share of a value
    among placebo subsets, locate an observed value among subsets and are not $p$-values.}
  \label{tab:intervals}
  \centering\footnotesize
  \begin{tabular}{@{}QRrS@{}}
    \toprule
    Study & Resampled & Replicates & Intervals \\
    \midrule
    WN18RR experiment & test triples; head synsets in the clustered sensitivity analysis &\val{2,000}{txt:scripts/analyze_wn18rr.py@'--reps', type=int, default=(\d+)} & percentile, per pair and estimand, unadjusted \\
    Panel size & tasks (BigCodeBench); questions within subject (MMLU) & 100 & percentile, per panel size \\
    Another prompt format & tasks & \val{2,000}{src:design constant, SEPARATED_SCORING_DESIGN.md} & pointwise for the curves of Fig.~\ref{fig:filters}b; for the excess penalty, simultaneous over the 20 contrasts of the curves \\
    Panel composition & tasks & \val{1,000}{src:design constant, PANEL_COMPOSITION_DESIGN.md Section 7} & percentile for $D(8)$; simultaneous over the eight line contrasts \\
    Penalty on the questions MMLU-Pro retained & questions within subject & \val{1,000}{src:design constant, MMLU_RECONSTRUCTION_DESIGN.md Section 4} & percentile \\
    Penalty on MMLU-Pro's own questions & questions within category and source & \val{1,000}{txt:MMLUPRO_FILTER_DESIGN.md@Question bootstrap\.\*\* ([\d,]+) replicates} & percentile \\
    Order on kept items & questions within subject (MMLU); instances (released subsets) & \val{2,000}{src:design constant, MMLU_RECONSTRUCTION_DESIGN.md Section 11 and SUBSET_C_DESIGN.md} & simultaneous over pairs \\
    \bottomrule
  \end{tabular}
\end{table}

\section*{Data availability}
\textit{Inputs used.} All inputs used are public. The benchmark releases are used as distributed: WN18RR and
YAGO3-10 with ConvE\citep{dettmers2018convolutional}, FB15k-237\citep{toutanova2015observed}, the Open Graph
Benchmark datasets\citep{hu2020open}, the Gowalla, Yelp2018 and Amazon-Book splits of
LightGCN\citep{he2020lightgcn}, and the TREC Deep Learning 2019 and 2020 passage
judgments\citep{craswell2020overview,craswell2021overview}. The declared universes that are available are the public
sources the benchmarks were built from, among them WordNet\citep{miller1995wordnet}~\val{3.0}{src:resource version, not a result}.
The filter studies use the public per-task and per-question results of the SWE-bench\citep{jimenez2024swebench},
BigCodeBench\citep{zhuo2025bigcodebench}, MMLU-Pro\citep{wang2024mmlupro} and Open LLM
Leaderboard\citep{beeching2023open} runs, the released annotations of SWE-bench
Verified\citep{chowdhury2024swebenchverified}, and the SWE-bench, BigCodeBench, MMLU-Pro and
MMLU\citep{hendrycks2021measuring} tables; the Open LLM Leaderboard runs come from its archive, \texttt{open-llm-leaderboard-old} on Hugging
Face. The reproduction package lists every input file with its public source and SHA-256: \texttt{inputs.json} for
the scope cards and the WN18RR study, and \texttt{filter\_inputs.json}, with the download manifests it names, for the
filter studies. Each scope card records the SHA-256 of the release files and a fingerprint of the universe's content.
\textit{Sources not available.} Three declared universes are not available: the 2013 Freebase snapshot behind
FB15k-237 is no longer distributed, ogbl-collab would need an external author graph, and Yelp2018's raw data are
behind Yelp's terms. These benchmarks are reported in regime (iii), with what is missing.
\textit{Outputs of this study.} The declared WN18RR target is not redistributed: scripts rebuild it from WordNet
and WN18RR and check it against recorded hashes. Nor are the inputs of the filter studies: \texttt{replay\_inputs.py}
downloads them again from their providers and checks each against the records of the original download. The
per-query scorer outputs of all runs, the stored outputs of the filter studies, among them the per-question
correctness matrix of the MMLU reconstruction, the records behind every reported value and the scope cards are
retained in the reproduction package (see Code availability).

\section*{Code availability}
ScopeCard (command line and browser page), the card library, the experiment scripts, \texttt{reproduce.py} and
\texttt{replay\_inputs.py} are maintained in a versioned reproduction package; this manuscript uses its release
v0.1.0, which contains ScopeCard v0.0.2. The package is at \url{https://github.com/JustinLi2002/scopecard}. The
browser page is served at \url{https://justinli2002.github.io/scopecard/app/}
and the card library at \url{https://justinli2002.github.io/scopecard/}.

\section*{Author contributions}
Justin Li conceived and conducted the study, developed the software, analysed the results and wrote the manuscript.

\section*{Competing interests}
The author declares no competing interests.

\bibliographystyle{unsrtnat}
\bibliography{references}

\clearpage
\section*{Extended Data}
\setcounter{table}{0}
\setcounter{figure}{0}
\captionsetup[table]{name=Extended Data Table}
\captionsetup[figure]{name=Extended Data Fig.}
\begin{table}[!ht]
  \caption{\textbf{Ten benchmarks.} By domain: evaluated unit, regime, decidability of the construction rule from
    published sources, $W$ at the unit, and the rule diagnosis: the Jaccard similarity $J$ of the stated rule and
    of the best simple rule found (threshold $T$) with the release.}
  \label{tab:nmi-ten}
  \centering\footnotesize
  \resizebox{\textwidth}{!}{%
\setlength{\tabcolsep}{3pt}
\begin{tabular}{>{\raggedright\arraybackslash}p{1.25in}>{\raggedright\arraybackslash}p{1.0in}>{\raggedright\arraybackslash}p{0.42in}>{\raggedright\arraybackslash}p{0.8in}>{\raggedright\arraybackslash}p{0.45in}>{\raggedright\arraybackslash}p{2.12in}}
\toprule
Benchmark & Unit & Regime & Decidability & $W$ & Rule diagnosis \\
\midrule
\multicolumn{6}{l}{\emph{Knowledge-graph link prediction}} \\
FB15k-237 & entity & (iii) & partial & — & — \\
WN18RR & entity & (i) & partial & 0.6336 & stated $J=0.149$; best both-ends $T=5$: $J=0.595$ \\
YAGO3-10 & entity & (i) & partial & 0.9540 & best both-ends $T=10$: $J=0.967$ \\
\addlinespace
\multicolumn{6}{l}{\emph{Graph link prediction}} \\
ogbl-collab & author pair & (iii) & undecidable & — & — \\
ogbl-citation2 & 2019 source paper & (i) & partial & 0.2025 & best threshold $T=6$: $J=1.000$ \\
\addlinespace
\multicolumn{6}{l}{\emph{Recommendation}} \\
Gowalla & user--item pair & (i) & partial & 0.9911 & best $k$-core $T=10$: $J=0.9999$ \\
Yelp2018 & user--item pair & (iii) & decidable, unverified & — & — \\
Amazon-Book & user--item pair & (i) & partial & 0.9997 & stated $J=0.502$; best $k$-core $T=14$: $J=0.726$ \\
\addlinespace
\multicolumn{6}{l}{\emph{Information retrieval}} \\
TREC DL 2019 passage & topic & (i) & verified & 0.7850 & stated rule: 0 violations, 43 judged topics \\
TREC DL 2020 passage & topic & (i) & verified & 0.7300 & stated rule: 0 violations, 54 judged topics \\
\bottomrule
\end{tabular}
\relax}
\end{table}

\begin{table}[!ht]
  \caption{\textbf{Validation of ScopeCard.} Times are wall-clock on a laptop CPU; details in Methods.}
  \label{ed:validation}
  \centering\small
  \begin{tabular}{@{}>{\raggedright\arraybackslash}p{1.1in}>{\raggedright\arraybackslash}p{5.0in}@{}}
    \toprule
    Check & Result \\
    \midrule
    V1 Pilot scripts
      & All pilot numbers of the \val{7}{src:tool/library/benchmarks.json, benchmarks with a card} benchmarks with
        a card and of WN18, from the same files (\val{10}{src:tool/tests/test_golden.py, number of golden tests}
        golden tests); WN18RR's mean $W_q$ equals the experiment's
        (\val{0.6337}{card:wn18rr@exclusion.per_query.W_bar},
        \val{6,266}{card:wn18rr@exclusion.per_query.queries} queries) \\
    V2 Known rules
      & \val{2,694}{txt:results/validation/v2_summary.md@\((\d+) used} runs, answers from the separate
        implementation: $W$ exact in all; pair-level $W$ equals enumeration in all
        \val{864}{txt:results/validation/v2_summary.md@unobserved pair in (\d+) of} small-graph runs; all
        \val{12}{txt:results/validation/v2_summary.md@- (\d+) runs; exit code 0} command-line runs exact \\
    V3 Refusals
      & \val{27}{txt:results/validation/v3_refusals.md@(\d+) of \d+ scenarios behave} of
        \val{27}{txt:results/validation/v3_refusals.md@\d+ of (\d+) scenarios behave} scenarios on the command
        line, \val{16}{txt:results/validation/v3_browser.md@(\d+) of \d+ rows behave} of
        \val{16}{txt:results/validation/v3_browser.md@\d+ of (\d+) rows behave} in the browser \\
    V4 Browser
      & Cards byte-identical for WN18RR, TREC DL 2019, Gowalla and a pair example; found two bugs \\
    V5 Reproduction
      & One command from public inputs and stored scorer outputs (no retraining); records, cards and tables match
        the stored copies in a new Python \val{3.12}{src:software version, not a result} environment
        (\val{5}{calc:{txt:results/validation/v5_clean_env/README.md@paper: (\d+) min}+{txt:results/validation/v5_clean_env/README.md@paper: \d+ min (\d+) s}/60}
        + \val{22}{calc:{txt:results/validation/v5_clean_env/README.md@cards: (\d+) min}+{txt:results/validation/v5_clean_env/README.md@cards: \d+ min (\d+) s}/60}
        min) \\
    V6 Users
      & Two internal tests, problems fixed; a test outside the project: pending \\
    V7 Cost
      & CPU only; \val{5}{txt:results/validation/v5_clean_env/README.md@wn18rr: card in (\d+) s} s (WN18RR) to
        \val{313}{txt:results/validation/v5_clean_env/README.md@amazon-book: card in (\d+) s} s (Amazon-Book);
        peak memory \val{2.9}{calc:{txt:results/validation/v7_memory/memory.tsv@yago3-10\t[\d.]+\t(\d+)}/1e9} GB
        (YAGO3-10) and \val{3.3}{calc:{txt:results/validation/v7_memory/memory.tsv@amazon-book\t[\d.]+\t(\d+)}/1e9} GB
        (Amazon-Book) \\
    \bottomrule
  \end{tabular}
\end{table}

\begin{table}[p]

  \caption{\textbf{The WN18RR scope card, abridged.} JSON and Markdown versions are in the supplementary code.}

  \label{ed:card}

  \small

  \begin{itemize}\setlength\itemsep{1pt}
\item \textbf{Scope statement:} Comparisons on WN18RR license claims about the entities it can evaluate, which
  leave out \val{63.4}{card:wn18rr@exclusion.W|pct}\% of the \val{112,199}{card:wn18rr@exclusion.units}
  WordNet~\val{3.0}{src:resource version, not a result} synsets that carry at least one triple of its 11
  relations. The closest simple rule (both ends, $T=5$) is not reproduced (Jaccard
  \val{0.595}{card:wn18rr@rule_diagnosis.best.jaccard}).
\item \textbf{Mapping:} \val{40,943}{card:wn18rr@mapping.mapped_ids} of
  \val{40,943}{card:wn18rr@mapping.release_ids} ids mapped through
  \val{92,998}{card:wn18rr@mapping.matched_triples} of \val{93,003}{card:wn18rr@mapping.release_triples}
  triples; \val{161}{card:wn18rr@mapping.merged_ids} ids merge synsets of different parts of speech.
\item \textbf{Exclusion:} $W = \val{0.6336}{card:wn18rr@exclusion.W}$
  (\val{71,094}{card:wn18rr@exclusion.excluded} of \val{112,199}{card:wn18rr@exclusion.units} synsets); by part of
  speech: adjectives \val{0.928}{card:wn18rr@exclusion.by_strata.pos.a.W}, nouns
  \val{0.608}{card:wn18rr@exclusion.by_strata.pos.n.W}, adverbs \val{0.898}{card:wn18rr@exclusion.by_strata.pos.r.W},
  verbs \val{0.432}{card:wn18rr@exclusion.by_strata.pos.v.W}; per filtered test query, mean
  $W_q = \val{0.6337}{card:wn18rr@exclusion.per_query.W_bar}$ over
  \val{6,266}{card:wn18rr@exclusion.per_query.queries} queries.
\item \textbf{Rule diagnosis:} stated (both ends, $T=15$): Jaccard
  \val{0.149}{card:wn18rr@rule_diagnosis.stated.jaccard}, not reproduced; best found (both ends, $T=5$):
  \val{0.595}{card:wn18rr@rule_diagnosis.best.jaccard}, not reproduced.
\item \textbf{Derived scope:} $d_{\min} = 3$.
\item \textbf{Hygiene:} \val{1,086}{card:wn18rr@hygiene.reverse_leakage.leaking} of
  \val{3,134}{card:wn18rr@hygiene.reverse_leakage.test_triples} test triples
  (\val{34.7}{card:wn18rr@hygiene.reverse_leakage.share|pct}\%) have their reverse in training, among them
  \val{1,011}{json:../tool/library/cards/wn18rr.json@hygiene.reverse_leakage.by_relation._derivationally_related_form.leaking}
  of the
  \val{1,074}{json:../tool/library/cards/wn18rr.json@hygiene.reverse_leakage.by_relation._derivationally_related_form.test}
  test triples of \texttt{derivationally\_related\_form};
  \val{209}{card:wn18rr@hygiene.unseen_test_units.unseen_units} test entities never occur in training
  (\val{210}{card:wn18rr@hygiene.unseen_test_units.test_triples_affected} test triples).
\end{itemize}

\end{table}

\begin{table}[p]
  \caption{V2: recovery of the generating rule when a share $\epsilon$ of the release's units is moved in or out
    at random (\val{389}{txt:results/validation/v2_summary.md@\| 0 \| (\d+) \|} runs per level). Recovered: the
    best rule found selects exactly the units of the generating rule. The verdict compares the best rule with
    the noisy release; $J$ is the Jaccard similarity of the generating rule with that release.}
  \label{ed:v2}
  \centering\small
  \begin{tabular}{@{}rrrrrrr@{}}
    \toprule
    $\epsilon$ & Recovered & Same family and $T$ & Reproduced & Approximately & Not reproduced & Mean $J$ \\
    \midrule
    \val{0}{src:design constant, scripts/validate_synthetic.py NOISE} & \val{100.0}{txt:results/validation/v2_summary.md@\| 0 \| \d+ \| ([\d.]+)}\% & \val{79.4}{txt:results/validation/v2_summary.md@\| 0 \| \d+ \| \S+ \| ([\d.]+)}\% & \val{100.0}{txt:results/validation/v2_summary.md@\| 0 \| \d+ \| \S+ \| \S+ \| ([\d.]+)}\% & \val{0.0}{txt:results/validation/v2_summary.md@\| 0 \| \d+ \| \S+ \| \S+ \| \S+ \| ([\d.]+)}\% & \val{0.0}{txt:results/validation/v2_summary.md@\| 0 \| \d+ \| \S+ \| \S+ \| \S+ \| \S+ \| ([\d.]+)}\% & \val{1.0000}{txt:results/validation/v2_summary.md@\| 0 \| \d+(?: \| \S+){5} \| ([\d.]+) \|} \\
    \val{0.005}{src:design constant, scripts/validate_synthetic.py NOISE} & \val{99.7}{txt:results/validation/v2_summary.md@\| 0\.005 \| \d+ \| ([\d.]+)}\% & \val{79.2}{txt:results/validation/v2_summary.md@\| 0\.005 \| \d+ \| \S+ \| ([\d.]+)}\% & \val{5.9}{txt:results/validation/v2_summary.md@\| 0\.005 \| \d+ \| \S+ \| \S+ \| ([\d.]+)}\% & \val{94.1}{txt:results/validation/v2_summary.md@\| 0\.005 \| \d+ \| \S+ \| \S+ \| \S+ \| ([\d.]+)}\% & \val{0.0}{txt:results/validation/v2_summary.md@\| 0\.005 \| \d+ \| \S+ \| \S+ \| \S+ \| \S+ \| ([\d.]+)}\% & \val{0.9903}{txt:results/validation/v2_summary.md@\| 0\.005 \| \d+(?: \| \S+){5} \| ([\d.]+) \|} \\
    \val{0.01}{src:design constant, scripts/validate_synthetic.py NOISE} & \val{100.0}{txt:results/validation/v2_summary.md@\| 0\.01 \| \d+ \| ([\d.]+)}\% & \val{79.4}{txt:results/validation/v2_summary.md@\| 0\.01 \| \d+ \| \S+ \| ([\d.]+)}\% & \val{2.6}{txt:results/validation/v2_summary.md@\| 0\.01 \| \d+ \| \S+ \| \S+ \| ([\d.]+)}\% & \val{97.4}{txt:results/validation/v2_summary.md@\| 0\.01 \| \d+ \| \S+ \| \S+ \| \S+ \| ([\d.]+)}\% & \val{0.0}{txt:results/validation/v2_summary.md@\| 0\.01 \| \d+ \| \S+ \| \S+ \| \S+ \| \S+ \| ([\d.]+)}\% & \val{0.9807}{txt:results/validation/v2_summary.md@\| 0\.01 \| \d+(?: \| \S+){5} \| ([\d.]+) \|} \\
    \val{0.02}{src:design constant, scripts/validate_synthetic.py NOISE} & \val{99.5}{txt:results/validation/v2_summary.md@\| 0\.02 \| \d+ \| ([\d.]+)}\% & \val{78.9}{txt:results/validation/v2_summary.md@\| 0\.02 \| \d+ \| \S+ \| ([\d.]+)}\% & \val{0.3}{txt:results/validation/v2_summary.md@\| 0\.02 \| \d+ \| \S+ \| \S+ \| ([\d.]+)}\% & \val{98.2}{txt:results/validation/v2_summary.md@\| 0\.02 \| \d+ \| \S+ \| \S+ \| \S+ \| ([\d.]+)}\% & \val{1.5}{txt:results/validation/v2_summary.md@\| 0\.02 \| \d+ \| \S+ \| \S+ \| \S+ \| \S+ \| ([\d.]+)}\% & \val{0.9604}{txt:results/validation/v2_summary.md@\| 0\.02 \| \d+(?: \| \S+){5} \| ([\d.]+) \|} \\
    \val{0.05}{src:design constant, scripts/validate_synthetic.py NOISE} & \val{91.3}{txt:results/validation/v2_summary.md@\| 0\.05 \| \d+ \| ([\d.]+)}\% & \val{70.7}{txt:results/validation/v2_summary.md@\| 0\.05 \| \d+ \| \S+ \| ([\d.]+)}\% & \val{0.0}{txt:results/validation/v2_summary.md@\| 0\.05 \| \d+ \| \S+ \| \S+ \| ([\d.]+)}\% & \val{0.0}{txt:results/validation/v2_summary.md@\| 0\.05 \| \d+ \| \S+ \| \S+ \| \S+ \| ([\d.]+)}\% & \val{100.0}{txt:results/validation/v2_summary.md@\| 0\.05 \| \d+ \| \S+ \| \S+ \| \S+ \| \S+ \| ([\d.]+)}\% & \val{0.9050}{txt:results/validation/v2_summary.md@\| 0\.05 \| \d+(?: \| \S+){5} \| ([\d.]+) \|} \\
    \val{0.1}{src:design constant, scripts/validate_synthetic.py NOISE} & \val{81.5}{txt:results/validation/v2_summary.md@\| 0\.1 \| \d+ \| ([\d.]+)}\% & \val{62.5}{txt:results/validation/v2_summary.md@\| 0\.1 \| \d+ \| \S+ \| ([\d.]+)}\% & \val{0.0}{txt:results/validation/v2_summary.md@\| 0\.1 \| \d+ \| \S+ \| \S+ \| ([\d.]+)}\% & \val{0.0}{txt:results/validation/v2_summary.md@\| 0\.1 \| \d+ \| \S+ \| \S+ \| \S+ \| ([\d.]+)}\% & \val{100.0}{txt:results/validation/v2_summary.md@\| 0\.1 \| \d+ \| \S+ \| \S+ \| \S+ \| \S+ \| ([\d.]+)}\% & \val{0.8195}{txt:results/validation/v2_summary.md@\| 0\.1 \| \d+(?: \| \S+){5} \| ([\d.]+) \|} \\
    \bottomrule
  \end{tabular}
\end{table}

\begin{table}[p]
  \caption{V3: the refusal and warning suite. Every scenario is a regression test.}
  \label{ed:v3}
  \centering\footnotesize
  \begin{tabular}{@{}l>{\raggedright\arraybackslash}p{2.75in}>{\raggedright\arraybackslash}p{3.2in}@{}}
    \toprule
    Id & Scenario & Expected behaviour \\
    \midrule
    R1 & Release ids do not match the universe (link prediction) & refused: fewer than half of the ids occur in the universe \\
    R2 & Release does not match the universe (retrieval) & refused: no judged topic is a released topic \\
    R3 & Release partly outside the universe & warned: at least half, but fewer than
      \val{95}{src:design constant, tool/src/scopecard/mapping.py MAP_WARN}\%, of the ids occur in it \\
    R4 & Malformed release line & refused, naming the file and the line \\
    R5 & One empty split in a link-prediction release & refused, naming the empty file \\
    R5b & One empty split in a LightGCN release & refused, naming the empty file \\
    R5c & All release files empty & refused, naming an empty file \\
    R5d & Empty universe unit list & refused, naming the empty file \\
    R5e & Pair file with a header only & refused, naming the empty file \\
    R5f & Empty topic list & refused, naming the empty file \\
    R5g & LightGCN split with user ids but no items & refused, naming the file \\
    R6 & Raw file with some lines of the wrong format & warned: skipped lines counted per file, the first one named \\
    R7 & Raw file with no line of the right format & refused \\
    R8 & Pair file with an unused label column & warned \\
    R9 & Sampled pairs with an unknown sampling rule & coverage only, no break-even \\
    R10 & Declared candidate rule that the file contradicts & warned, and the break-even withheld \\
    R11 & Declared candidate rule that the file supports (control) & no warning; the exclusion is structural \\
    R12 & Break-even asked for a metric that does not decompose (MRR) & withheld \\
    R13 & Model difference given without its metric & refused by the argument parser \\
    R14 & No unobserved pair in the universe & pair-level $W$ undefined, with a note \\
    R15 & Nothing excluded ($W=0$), with a model difference & the comparison states that nothing excluded can
      reverse it \\
    R16 & Two unit sets that share ids (bipartite) & no warning: equal ids on the two sides are distinct units \\
    R17 & Recommendation release counted with the pilot table's convention & the share is the headline; the pilot
      value is kept beside it \\
    R18 & Corpus smaller than the documents judged for one topic & refused, naming the corpus size \\
    R19 & Model difference that is not finite or not in $[-1,1]$ & refused as an input error, also in the page \\
    R20 & Model difference while some test triples fall outside the universe & warned: the break-even uses fewer
      queries than the difference and carries no guarantee \\
    R21 & A test query with no kept candidate ($W_q = 1$) & break-even withheld: the benchmark difference is
      undefined on that query \\
    \bottomrule
  \end{tabular}
\end{table}

\begin{table}[p]
  \caption{Query side: MRR over all candidates on the held-out triples with an end in $U$, by the regions of the
    query and answer synsets, range over seeds. The last two columns split the queries by whether their reverse is
    an E2 training triple (post hoc).}
  \label{ed:query-side}
  \centering\footnotesize
  \begin{tabular}{@{}lllllll@{}}
    \toprule
    & All & $O\to U$ & $U\to O$ & $U\to U$ & Leaking & Not leaking \\
    Model & (\val{6,268}{v:E2/query_side:all/complex/0/rr_C.n}) & (\val{2,184}{v:E2/query_side:O->U/complex/0/rr_C.n}) & (\val{2,184}{v:E2/query_side:U->O/complex/0/rr_C.n}) & (\val{1,900}{v:E2/query_side:U->U/complex/0/rr_C.n}) & (\val{3,568}{v:E2/posthoc_u_test:leak/complex/0/rr_C.n}) & (\val{2,700}{v:E2/posthoc_u_test:no_leak/complex/0/rr_C.n}) \\
    \midrule
    Popularity & \val{0.014}{v:E2/query_side:all/popularity/0/rr_C} & \val{0.0001}{v:E2/query_side:O->U/popularity/0/rr_C} & \val{0.036}{v:E2/query_side:U->O/popularity/0/rr_C} & \val{0.005}{v:E2/query_side:U->U/popularity/0/rr_C} & \val{0.006}{v:E2/posthoc_u_test:leak/popularity/0/rr_C} & \val{0.024}{v:E2/posthoc_u_test:no_leak/popularity/0/rr_C} \\
    TransE & \val{0.304}{min:E2/query_side:all/transe/0,1,2,3,4/rr_C}\textendash{}\val{0.306}{max:E2/query_side:all/transe/0,1,2,3,4/rr_C} & \val{0.163}{min:E2/query_side:O->U/transe/0,1,2,3,4/rr_C}\textendash{}\val{0.168}{max:E2/query_side:O->U/transe/0,1,2,3,4/rr_C} & \val{0.351}{min:E2/query_side:U->O/transe/0,1,2,3,4/rr_C}\textendash{}\val{0.356}{max:E2/query_side:U->O/transe/0,1,2,3,4/rr_C} & \val{0.406}{min:E2/query_side:U->U/transe/0,1,2,3,4/rr_C}\textendash{}\val{0.412}{max:E2/query_side:U->U/transe/0,1,2,3,4/rr_C} & \val{0.432}{min:E2/posthoc_u_test:leak/transe/0,1,2,3,4/rr_C}\textendash{}\val{0.436}{max:E2/posthoc_u_test:leak/transe/0,1,2,3,4/rr_C} & \val{0.132}{min:E2/posthoc_u_test:no_leak/transe/0,1,2,3,4/rr_C}\textendash{}\val{0.135}{max:E2/posthoc_u_test:no_leak/transe/0,1,2,3,4/rr_C} \\
    ConvE & \val{0.618}{min:E2/query_side:all/conve/0,1,2,3,4/rr_C}\textendash{}\val{0.620}{max:E2/query_side:all/conve/0,1,2,3,4/rr_C} & \val{0.476}{min:E2/query_side:O->U/conve/0,1,2,3,4/rr_C}\textendash{}\val{0.481}{max:E2/query_side:O->U/conve/0,1,2,3,4/rr_C} & \val{0.557}{min:E2/query_side:U->O/conve/0,1,2,3,4/rr_C}\textendash{}\val{0.564}{max:E2/query_side:U->O/conve/0,1,2,3,4/rr_C} & \val{0.846}{min:E2/query_side:U->U/conve/0,1,2,3,4/rr_C}\textendash{}\val{0.848}{max:E2/query_side:U->U/conve/0,1,2,3,4/rr_C} & \val{0.995}{min:E2/posthoc_u_test:leak/conve/0,1,2,3,4/rr_C}\textendash{}\val{0.998}{max:E2/posthoc_u_test:leak/conve/0,1,2,3,4/rr_C} & \val{0.117}{min:E2/posthoc_u_test:no_leak/conve/0,1,2,3,4/rr_C}\textendash{}\val{0.123}{max:E2/posthoc_u_test:no_leak/conve/0,1,2,3,4/rr_C} \\
    DistMult & \val{0.634}{min:E2/query_side:all/distmult/0,1,2,3,4/rr_C}\textendash{}\val{0.636}{max:E2/query_side:all/distmult/0,1,2,3,4/rr_C} & \val{0.489}{min:E2/query_side:O->U/distmult/0,1,2,3,4/rr_C}\textendash{}\val{0.490}{max:E2/query_side:O->U/distmult/0,1,2,3,4/rr_C} & \val{0.588}{min:E2/query_side:U->O/distmult/0,1,2,3,4/rr_C}\textendash{}\val{0.593}{max:E2/query_side:U->O/distmult/0,1,2,3,4/rr_C} & \val{0.853}{min:E2/query_side:U->U/distmult/0,1,2,3,4/rr_C}\textendash{}\val{0.854}{max:E2/query_side:U->U/distmult/0,1,2,3,4/rr_C} & \val{0.999}{min:E2/posthoc_u_test:leak/distmult/0,1,2,3,4/rr_C} & \val{0.151}{min:E2/posthoc_u_test:no_leak/distmult/0,1,2,3,4/rr_C}\textendash{}\val{0.156}{max:E2/posthoc_u_test:no_leak/distmult/0,1,2,3,4/rr_C} \\
    RESCAL & \val{0.630}{min:E2/query_side:all/rescal/0,1,2,3,4/rr_C}\textendash{}\val{0.631}{max:E2/query_side:all/rescal/0,1,2,3,4/rr_C} & \val{0.495}{min:E2/query_side:O->U/rescal/0,1,2,3,4/rr_C}\textendash{}\val{0.496}{max:E2/query_side:O->U/rescal/0,1,2,3,4/rr_C} & \val{0.575}{min:E2/query_side:U->O/rescal/0,1,2,3,4/rr_C}\textendash{}\val{0.577}{max:E2/query_side:U->O/rescal/0,1,2,3,4/rr_C} & \val{0.849}{min:E2/query_side:U->U/rescal/0,1,2,3,4/rr_C}\textendash{}\val{0.850}{max:E2/query_side:U->U/rescal/0,1,2,3,4/rr_C} & \val{0.998}{min:E2/posthoc_u_test:leak/rescal/0,1,2,3,4/rr_C}\textendash{}\val{0.999}{max:E2/posthoc_u_test:leak/rescal/0,1,2,3,4/rr_C} & \val{0.144}{min:E2/posthoc_u_test:no_leak/rescal/0,1,2,3,4/rr_C}\textendash{}\val{0.146}{max:E2/posthoc_u_test:no_leak/rescal/0,1,2,3,4/rr_C} \\
    ComplEx & \val{0.637}{min:E2/query_side:all/complex/0,1,2,3,4/rr_C}\textendash{}\val{0.640}{max:E2/query_side:all/complex/0,1,2,3,4/rr_C} & \val{0.499}{min:E2/query_side:O->U/complex/0,1,2,3,4/rr_C}\textendash{}\val{0.506}{max:E2/query_side:O->U/complex/0,1,2,3,4/rr_C} & \val{0.581}{min:E2/query_side:U->O/complex/0,1,2,3,4/rr_C}\textendash{}\val{0.587}{max:E2/query_side:U->O/complex/0,1,2,3,4/rr_C} & \val{0.853}{min:E2/query_side:U->U/complex/0,1,2,3,4/rr_C}\textendash{}\val{0.857}{max:E2/query_side:U->U/complex/0,1,2,3,4/rr_C} & \val{0.998}{min:E2/posthoc_u_test:leak/complex/0,1,2,3,4/rr_C}\textendash{}\val{0.999}{max:E2/posthoc_u_test:leak/complex/0,1,2,3,4/rr_C} & \val{0.159}{min:E2/posthoc_u_test:no_leak/complex/0,1,2,3,4/rr_C}\textendash{}\val{0.165}{max:E2/posthoc_u_test:no_leak/complex/0,1,2,3,4/rr_C} \\
    \bottomrule
  \end{tabular}
\end{table}

\begin{table}[p]
  \caption{Training/protocol effect on $O$: E2 minus E1 for the same model and seed, range over seeds
    (popularity is deterministic and has one run).}
  \label{ed:protocol}
  \centering\small
  \begin{tabular}{@{}lrll@{}}
    \toprule
    Model & Runs & AUC($O$) & MRR($O$) \\
    \midrule
    Popularity & 1 & \val{+0.0054}{v:E1/training_side/popularity/0/train_effect_auc_O} & \val{-0.0036}{v:E1/training_side/popularity/0/train_effect_mrr_O} \\
    TransE & 5 & \val{+0.0144}{min:E1/training_side/transe/0,1,2,3,4/train_effect_auc_O} to \val{+0.0174}{max:E1/training_side/transe/0,1,2,3,4/train_effect_auc_O} & \val{+0.0085}{min:E1/training_side/transe/0,1,2,3,4/train_effect_mrr_O} to \val{+0.0114}{max:E1/training_side/transe/0,1,2,3,4/train_effect_mrr_O} \\
    ConvE & 5 & \val{+0.0087}{min:E1/training_side/conve/0,1,2,3,4/train_effect_auc_O} to \val{+0.0139}{max:E1/training_side/conve/0,1,2,3,4/train_effect_auc_O} & \val{-0.0005}{min:E1/training_side/conve/0,1,2,3,4/train_effect_mrr_O} to \val{+0.0048}{max:E1/training_side/conve/0,1,2,3,4/train_effect_mrr_O} \\
    DistMult & 5 & \val{+0.0092}{min:E1/training_side/distmult/0,1,2,3,4/train_effect_auc_O} to \val{+0.0160}{max:E1/training_side/distmult/0,1,2,3,4/train_effect_auc_O} & \val{+0.0025}{min:E1/training_side/distmult/0,1,2,3,4/train_effect_mrr_O} to \val{+0.0098}{max:E1/training_side/distmult/0,1,2,3,4/train_effect_mrr_O} \\
    RESCAL & 5 & \val{+0.0014}{min:E1/training_side/rescal/0,1,2,3,4/train_effect_auc_O} to \val{+0.0118}{max:E1/training_side/rescal/0,1,2,3,4/train_effect_auc_O} & \val{+0.00002}{min:E1/training_side/rescal/0,1,2,3,4/train_effect_mrr_O} to \val{+0.0030}{max:E1/training_side/rescal/0,1,2,3,4/train_effect_mrr_O} \\
    ComplEx & 5 & \val{+0.0192}{min:E1/training_side/complex/0,1,2,3,4/train_effect_auc_O} to \val{+0.0246}{max:E1/training_side/complex/0,1,2,3,4/train_effect_auc_O} & \val{+0.0061}{min:E1/training_side/complex/0,1,2,3,4/train_effect_mrr_O} to \val{+0.0181}{max:E1/training_side/complex/0,1,2,3,4/train_effect_mrr_O} \\
    \bottomrule
  \end{tabular}
\end{table}

\begin{table}[p]
  \caption{NBFNet deployment test per seed (primary estimand) and its graph-side effect on $O$. Intervals for
    the seed mean are in Extended Data Table~\ref{ed:models}.}
  \label{ed:deploy}
  \centering\footnotesize
  \begin{tabular}{@{}lrrrrrrrr@{}}
    \toprule
    & & AUC($U$) & & MRR($C$) & Outranked & Top-10 & \multicolumn{2}{c}{Graph side} \\
    Seed & AUC($O$) & $-$ AUC($O$) & MRR($O$) & $-$ MRR($O$) & by $U$ & drop & AUC($O$) & MRR($O$) \\
    \midrule
    0 & \val{0.9852}{v:deploy/primary/nbfnet/0/auc_O} & \val{+0.0047}{v:deploy/primary/nbfnet/0/auc_U_minus_O} & \val{0.5655}{v:deploy/primary/nbfnet/0/rr_O} & \val{-0.0184}{v:deploy/primary/nbfnet/0/mrr_C_minus_O} & \val{0.364}{v:deploy/primary/nbfnet/0/intrusion} & \val{0.019}{v:deploy/primary/nbfnet/0/top10_drop} & \val{+0.0013}{v:deploy/graph_side/nbfnet/0/graph_effect_auc_O} & \val{+0.0207}{v:deploy/graph_side/nbfnet/0/graph_effect_mrr_O} \\
    1 & \val{0.9864}{v:deploy/primary/nbfnet/1/auc_O} & \val{+0.0001}{v:deploy/primary/nbfnet/1/auc_U_minus_O} & \val{0.5568}{v:deploy/primary/nbfnet/1/rr_O} & \val{-0.0217}{v:deploy/primary/nbfnet/1/mrr_C_minus_O} & \val{0.389}{v:deploy/primary/nbfnet/1/intrusion} & \val{0.028}{v:deploy/primary/nbfnet/1/top10_drop} & \val{+0.0021}{v:deploy/graph_side/nbfnet/1/graph_effect_auc_O} & \val{+0.0110}{v:deploy/graph_side/nbfnet/1/graph_effect_mrr_O} \\
    2 & \val{0.9871}{v:deploy/primary/nbfnet/2/auc_O} & \val{+0.0003}{v:deploy/primary/nbfnet/2/auc_U_minus_O} & \val{0.5732}{v:deploy/primary/nbfnet/2/rr_O} & \val{-0.0204}{v:deploy/primary/nbfnet/2/mrr_C_minus_O} & \val{0.367}{v:deploy/primary/nbfnet/2/intrusion} & \val{0.027}{v:deploy/primary/nbfnet/2/top10_drop} & \val{+0.0025}{v:deploy/graph_side/nbfnet/2/graph_effect_auc_O} & \val{+0.0256}{v:deploy/graph_side/nbfnet/2/graph_effect_mrr_O} \\
    3 & \val{0.9869}{v:deploy/primary/nbfnet/3/auc_O} & \val{-0.0027}{v:deploy/primary/nbfnet/3/auc_U_minus_O} & \val{0.5678}{v:deploy/primary/nbfnet/3/rr_O} & \val{-0.0219}{v:deploy/primary/nbfnet/3/mrr_C_minus_O} & \val{0.380}{v:deploy/primary/nbfnet/3/intrusion} & \val{0.033}{v:deploy/primary/nbfnet/3/top10_drop} & \val{+0.0028}{v:deploy/graph_side/nbfnet/3/graph_effect_auc_O} & \val{+0.0218}{v:deploy/graph_side/nbfnet/3/graph_effect_mrr_O} \\
    4 & \val{0.9865}{v:deploy/primary/nbfnet/4/auc_O} & \val{+0.0035}{v:deploy/primary/nbfnet/4/auc_U_minus_O} & \val{0.5709}{v:deploy/primary/nbfnet/4/rr_O} & \val{-0.0200}{v:deploy/primary/nbfnet/4/mrr_C_minus_O} & \val{0.356}{v:deploy/primary/nbfnet/4/intrusion} & \val{0.021}{v:deploy/primary/nbfnet/4/top10_drop} & \val{+0.0018}{v:deploy/graph_side/nbfnet/4/graph_effect_auc_O} & \val{+0.0240}{v:deploy/graph_side/nbfnet/4/graph_effect_mrr_O} \\
    mean & \val{0.9864}{v:deploy/primary/nbfnet/mean/auc_O} & \val{+0.0012}{v:deploy/primary/nbfnet/mean/auc_U_minus_O} & \val{0.5668}{v:deploy/primary/nbfnet/mean/rr_O} & \val{-0.0205}{v:deploy/primary/nbfnet/mean/mrr_C_minus_O} & \val{0.371}{v:deploy/primary/nbfnet/mean/intrusion} & \val{0.026}{v:deploy/primary/nbfnet/mean/top10_drop} & --- & --- \\
    \bottomrule
  \end{tabular}
\end{table}

\begin{sidewaystable}
  \caption{Per model on the declared target: primary estimand, seed means with
    \val{95}{src:design constant, MEASURED_C_DESIGN.md Sect. 8}\% triple-bootstrap intervals. The NBFNet row is the
    deployment test (Methods), a separate estimand.}
  \label{ed:models}
  \centering\footnotesize
  \resizebox{\textheight}{!}{%
\begin{tabular}{lllllll}
\toprule
Model & W & AUC(O) & AUC(U) $-$ AUC(O) & MRR(O) & MRR(C) $-$ MRR(O) & Outranked by U \\
\midrule
Popularity & 0.6337 & 0.6204 [0.6142, 0.6260] & +0.0811 [+0.0790, +0.0832] & 0.0219 [0.0191, 0.0247] & $-$0.0001 [$-$0.0001, $-$0.00005] & 0.800 [0.791, 0.808] \\
TransE & 0.6337 & 0.9702 [0.9667, 0.9737] & +0.0033 [+0.0025, +0.0041] & 0.2379 [0.2302, 0.2457] & $-$0.0134 [$-$0.0146, $-$0.0122] & 0.534 [0.520, 0.547] \\
ConvE & 0.6337 & 0.8980 [0.8922, 0.9044] & +0.0211 [+0.0199, +0.0223] & 0.4434 [0.4273, 0.4596] & $-$0.0060 [$-$0.0068, $-$0.0053] & 0.515 [0.499, 0.530] \\
DistMult & 0.6337 & 0.9037 [0.8981, 0.9095] & $-$0.0042 [$-$0.0049, $-$0.0036] & 0.4580 [0.4425, 0.4737] & $-$0.0119 [$-$0.0133, $-$0.0104] & 0.506 [0.490, 0.522] \\
RESCAL & 0.6337 & 0.8762 [0.8695, 0.8833] & $-$0.0180 [$-$0.0191, $-$0.0168] & 0.4694 [0.4537, 0.4847] & $-$0.0071 [$-$0.0080, $-$0.0062] & 0.488 [0.473, 0.503] \\
ComplEx & 0.6337 & 0.9179 [0.9118, 0.9238] & +0.0024 [+0.0022, +0.0026] & 0.4859 [0.4705, 0.5010] & $-$0.0101 [$-$0.0113, $-$0.0091] & 0.483 [0.468, 0.498] \\
NBFNet (deployment test) & 0.6337 & 0.9864 [0.9848, 0.9880] & +0.0012 [+0.0003, +0.0021] & 0.5668 [0.5526, 0.5824] & $-$0.0205 [$-$0.0221, $-$0.0189] & 0.371 [0.357, 0.385] \\
\bottomrule
\end{tabular}
\relax}
\end{sidewaystable}

\begin{table}[p]
  \caption{GPU time of the WN18RR experiment. Cluster tasks include training, testing and scoring; the laptop
    time covers training and LibKGE's test, not the per-query scoring done afterwards.}
  \label{ed:compute}
  \centering\small
  \begin{tabular}{@{}llrr@{}}
    \toprule
    Design & GPU & Runs & GPU-hours \\
    \midrule
    E1, seed 0 (embedding models) & laptop & \val{5}{txt:results/compute/gpu_hours.md@\| E1 seed 0 \| laptop \| (\d+) \|} & \val{9.83}{txt:results/compute/gpu_hours.md@\| E1 seed 0 \| laptop \| \d+ \| ([\d.]+) \|} \\
    E1, seeds 1--4 (embedding models) & cluster & \val{20}{txt:results/compute/gpu_hours.md@\| E1 seeds 1-4 \| cluster \| (\d+) \|} & \val{13.92}{txt:results/compute/gpu_hours.md@\| E1 seeds 1-4 \| cluster \| \d+ \| ([\d.]+) \|} \\
    E2 & cluster & \val{25}{txt:results/compute/gpu_hours.md@\| E2 \| cluster \| (\d+) \|} & \val{57.27}{txt:results/compute/gpu_hours.md@\| E2 \| cluster \| \d+ \| ([\d.]+) \|} \\
    S3 & cluster & \val{25}{txt:results/compute/gpu_hours.md@\| S3 \| cluster \| (\d+) \|} & \val{56.77}{txt:results/compute/gpu_hours.md@\| S3 \| cluster \| \d+ \| ([\d.]+) \|} \\
    NBFNet: E1 training and the deployment test & cluster & \val{5}{txt:results/compute/gpu_hours.md@\| NBFNet E1 and deployment \| cluster \| (\d+) \|} & \val{46.69}{txt:results/compute/gpu_hours.md@\| NBFNet E1 and deployment \| cluster \| \d+ \| ([\d.]+) \|} \\
    NBFNet feasibility run & cluster & \val{1}{txt:results/compute/gpu_hours.md@\| NBFNet feasibility \| cluster \| (\d+) \|} & \val{3.00}{txt:results/compute/gpu_hours.md@\| NBFNet feasibility \| cluster \| \d+ \| ([\d.]+) \|} \\
    \midrule
    Total & & \val{81}{txt:results/compute/gpu_hours.md@\| total \| both \| (\d+) \|} & \val{187.48}{txt:results/compute/gpu_hours.md@\| total \| both \| \d+ \| ([\d.]+) \|} \\
    \bottomrule
  \end{tabular}
\end{table}

\begin{figure}[p]
  \centering
  \includegraphics[width=0.6\textwidth]{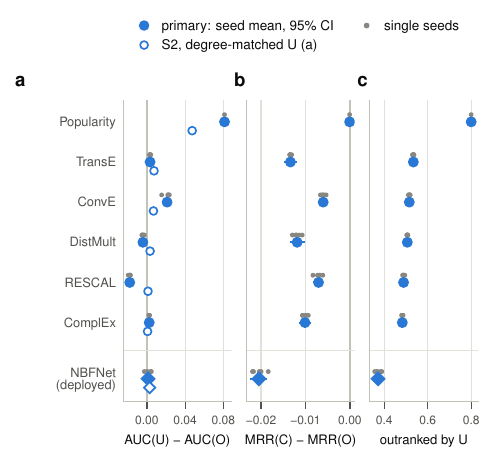}
  \caption{\textbf{Admitting the excluded synsets, per model} (E2, primary estimand unless marked). \textbf{a}, AUC
    on $U$ minus AUC on $O$, primary and degree-matched (S2). \textbf{b}, MRR on all candidates minus MRR on $O$.
    \textbf{c}, Share of queries whose positive is outranked by at least one excluded synset. Seed means with
    \val{95}{src:design constant, MEASURED_C_DESIGN.md Sect. 8}\% triple-bootstrap intervals; single seeds as grey
    dots. The NBFNet row (diamonds) is the deployment test, a separate estimand.}
  \label{fig:nmi-models}
\end{figure}

\begin{table}[p]
  \caption{\textbf{What each conclusion rests on.} Inputs, checks and what carries each conclusion: a field of the
    current scope card, the scripts of the reproduction package, or a reporting field recommended for releases of
    benchmarks filtered by model outputs (\emph{recommended}), which the current scope card does not record.}
  \label{ed:claims}
  \centering\footnotesize
  \begin{tabular}{@{}QRST@{}}
    \toprule
    Conclusion & Inputs & Checks & Carried by \\
    \midrule
    Regime, $W_q$, $c^*$ & release, declared universe, protocol & a separate plain implementation; synthetic releases;
      refusal scenarios (Extended Data Table~\ref{ed:validation}) & scope card: regime, exclusion, $c^*$ \\
    Rule diagnosis & release, universe & synthetic releases of every rule family, with and without noise & scope
      card: rule diagnosis \\
    WN18RR measurement & target graph; per-query scores of trained models & prespecified design; published values
      reproduced; one command rebuilds the analyses & scripts (a study) \\
    Own-vote identity & item-level outcomes of the panel & algebra (Methods); synthetic matrices & scripts;
      \emph{recommended:} whether scored outputs voted \\
    Panel-size relation & item-level outcomes; random panels & task bootstrap; placebo panels & scripts;
      \emph{recommended:} panel members and size \\
    Reach to another prompt format & outcomes in two prompt formats & synthetic checks; explicit resampling & scripts;
      \emph{recommended:} which outputs are scored \\
    Reach to a member's line & outcomes; model lines read from names & constant, independent and planted scoring
      vectors; null calibration of the intervals & scripts; \emph{recommended:} members' lines \\
    Penalty on the questions MMLU-Pro retained & MMLU-Pro's released questions; archived MMLU runs & fidelity
      gates; placebo pool & scripts; \emph{recommended:} rule and tie handling \\
    Order against references & item-level outcomes, subjects, answer keys & reproduction of the earlier reference;
      synthetic matrices; sensitivity analyses & scripts (a study) \\
    \bottomrule
  \end{tabular}
\end{table}

\begin{table}[p]
  \caption{\textbf{Notation.} The symbols of the Results and Methods, by part. Symbols local to one proof ($x$ and $y$
    in the bound on reciprocal rank) are not listed.}
  \label{ed:notation}
  \centering\footnotesize
  \begin{tabular}{@{}>{\raggedright\arraybackslash}p{9em}>{\raggedright\arraybackslash}p{\dimexpr\textwidth-9em-2\tabcolsep\relax}@{}}
    \toprule
    Symbol & Meaning \\
    \midrule
    \multicolumn{2}{@{}l}{\emph{Scope and identification}} \\
    $V$ & declared universe: the units a claim is meant to cover, at the unit the benchmark evaluates \\
    $O$, $U$ & the part a release keeps, $O\subseteq V$, and the excluded part, $U = V\setminus O$ \\
    $W$ & excluded share at the evaluated unit; where exclusion removes whole queries, the share of the queries \\
    $Q$, $q$; $Q_O$, $Q_U$ & the test queries and a query; the kept and the excluded queries when exclusion removes whole
      queries \\
    $p_q$, $C_q$ & the positive of query $q$, and its candidates: the units of $V$ it ranks, filtered of other known
      positives \\
    $O_q$, $U_q$ & kept and excluded candidates of $q$, $C_q\cap O$ and $C_q\cap U$ \\
    $W_q$, $\bar W$ & excluded share of the candidates of $q$, $|U_q|/|C_q|$, and its mean over the test queries \\
    $A$, $A'$, $B$ & models compared \\
    $\Delta_O$, $\Delta_U$, $\Delta_{\mathrm{target}}$ & two models' mean differences over the kept, the excluded and all
      queries, or in per-query AUC on $O_q$, $U_q$ and $C_q$; $\Delta_{O,q}$, $\Delta_{U,q}$ and $\Delta_{C,q}$ are the
      differences on one query \\
    $c_q$, $\bar c$, $c$ & change between the regions, $c_q = \Delta_{O,q} - \Delta_{U,q}$; its mean,
      $\Delta_O - \Delta_U$; its $W_q$-weighted mean \\
    $c^*$ & break-even change, $|\Delta_O|/\bar W$ \\
    $\operatorname{Cov}_q$ & covariance over the test queries, an unweighted population moment \\
    $k$ & cut-off of Hits@$k$ \\
    $S$, $s$, $v$ & a candidate set, a scoring function and a candidate \\
    $b_q(S)$, $a_q(S)$ & candidates in $S$ scored below the positive of $q$, ties counting one half; candidates in $S$
      scored above it \\
    $R$, $\pi$ & a release (Remark~\ref{rem:nmi-absence}) and a sampler's inclusion probability \\
    \addlinespace
    \multicolumn{2}{@{}l}{\emph{ScopeCard}} \\
    $T$ & threshold of a depth rule in the rule diagnosis \\
    $J$ & Jaccard similarity of a rule's units with the release \\
    $d_{\min}$ & derived scope: the smallest depth at which at most half of the units are excluded \\
    $\epsilon$ & share of units moved at random in the synthetic validation \\
    \addlinespace
    \multicolumn{2}{@{}l}{\emph{Filters that use model outputs}} \\
    $P$, $P'$ & a panel, and the panel with member $m$ replaced by $u$ \\
    $m$, $u$, $j$ & a panel member, a model outside the panel, any model \\
    $i$, $y_{ji}$, $y_m$ & an item; the outcome (0 or 1) of model $j$ on item $i$ in the outputs that vote; the outcomes
      of $m$ over the items \\
    $H$, $H'$ & kept sets of $P$ and $P'$ \\
    $E$, $L$; $e$, $l$ & items kept only when $m$ votes, $H\setminus H'$, and only when $u$ votes, $H'\setminus H$; their
      numbers \\
    $z$, $\bar z_X$ & a scoring vector over the items, and its mean over a set $X$ of items \\
    $\ell$, $n$ & focal line and number of builders in the panel composition study \\
    $w$, $d_w$ & an evaluation model, and its score on the spread panel's kept set minus its score on the concentrated
      panel's \\
    $D(\ell,n)$, $D(8)$ & mean concentration cost of the evaluation models of line $\ell$ at $n$ builders; its mean over
      the seven focal lines at 8 builders \\
    $G_m$ & controls of filter model $m$ \\
    $\alpha_m(\cdot)$, $\bar\alpha_{G_m}(\cdot)$ & accuracy of $m$ on a set of questions, and the mean accuracy of its
      controls \\
    $\delta_m$, $\bar\delta$ & penalty of filter model $m$ on the questions MMLU-Pro retained, and its mean over the five
      filter models \\
    R0, R1, R2 & reference subsets: random; matched on estimated difficulty; matched also on the answer key \\
    $|t|$ & absolute studentized bootstrap statistic; simultaneous intervals use its maximum \\
    \bottomrule
  \end{tabular}
\end{table}

\end{document}